\documentclass[preprint,review,5p,times,twocolumn,authoryear]{elsarticle}
\usepackage[ruled,vlined,linesnumbered]{algorithm2e}
\usepackage{array}
\usepackage[caption=false]{subfig}
\usepackage{stfloats}
\usepackage{url}
\usepackage{verbatim}
\usepackage{amsmath,amssymb,amsfonts}
\usepackage{graphicx}
\usepackage{textcomp}
\usepackage{xcolor}
\usepackage{booktabs}
\usepackage{hyperref}
\usepackage{pifont}
\usepackage{tabularray}
\usepackage{comment}
\usepackage{multicol}
\usepackage{multirow}
\usepackage{tabularx}
\usepackage{tikz} 
\usepackage{float}
\newcommand*\circled[1]{\tikz[baseline=(char.base)]{
            \node[shape=circle,draw,inner sep=1pt] (char) {#1};}}
            
\hypersetup{
  colorlinks   = true, %Colours links instead of ugly boxes
  urlcolor     = blue, %Colour for external hyperlinks
  linkcolor    = blue, %Colour of internal links
  citecolor   = blue %Colour of citations
}

\usepackage{enumitem}
\setlist{  
  listparindent=\parindent,
  parsep=0pt,
}

\begin{document}

\title{A Privacy-Preserving Multi-Stage Fall Detection Framework with Semi-supervised Federated Learning and Robotic Vision Confirmation}

\author[1]{Seyed Alireza Rahimi Azghadi}
\author[1]{Truong-Thanh-Hung Nguyen}
\author[3]{Hélène Fournier}
\author[4,1]{Monica Wachowicz}
\author[3,1]{René Richard}
\author[2]{Francis Palma}
\author[1]{Hung Cao}

\address[1]{Analytics Everywhere Lab, University of New Brunswick, Canada}
\address[2]{SE+AI Research Lab, University of New Brunswick, Canada}
\address[3]{National Research Council, New Brunswick, Canada}
\address[4]{RMIT University, Australia}

\begin{abstract}
The aging population is growing rapidly, and so is the danger of falls in older adults. A major cause of injury is falling, and detection in time can greatly save medical expenses and recovery time. However, to provide timely intervention and avoid unnecessary alarms, detection systems must be effective and reliable while addressing privacy concerns regarding the user. In this work, we propose a framework for detecting falls using several complementary systems: a semi-supervised federated learning-based fall detection system (SF2D), an indoor localization and navigation system, and a vision-based human fall recognition system. A wearable device and an edge device identify a fall scenario in the first system. On top of that, the second system uses an indoor localization technique first to localize the fall location and then navigate a robot to inspect the scenario. A vision-based detection system running on an edge device with a mounted camera on a robot is used to recognize fallen people. Each of the systems of this proposed framework achieves different accuracy rates. Specifically, the SF2D has a 0.81\% failure rate equivalent to 99.19\% accuracy, while the vision-based fallen people detection achieves 96.3\% accuracy. However, when we combine the accuracy of these two systems with the accuracy of the navigation system (95\% success rate), our proposed framework creates a highly reliable performance for fall detection, with an overall accuracy of 99.99\%. Not only is the proposed framework safe for older adults, but it is also a privacy-preserving solution for detecting falls.
\end{abstract}


\nonumnote{This work was supported by the NSERC Discovery Grant (NSERC RGPIN 2025-00129) \& NSERC Discovery Launch Supplement Fund (NSERC DGECR 2025-00129). The equipment used in the experiments was supported by CFI Project Number 39473 - Smart Campus Integration and Testing (SCIT Lab). The Cloud resource is supported by The Digital Research Alliance of Canada.}

\maketitle

\section{Introduction}
Advancements in public health have led to decreased mortality rates, and together with declining fertility rates, these factors are the main contributors to the aging global population. Although population aging is a global phenomenon, life expectancy among older adults varies based on socioeconomic status, health, and region. A significant portion of the global population can anticipate living into their 60s and beyond. In Brazil, for example, a child can expect to live up to 20 years longer than just 50 years ago. A similar trajectory is also occurring in other countries~\citep{schmidt2024future}. The rapid aging of populations around the world has several implications, such as rising healthcare demand and costs, a greater need for social services, a shrinking workforce, and innovation opportunities. One key implication of the rapid aging situation is that, in older age, those with the greatest health needs often have the fewest resources available to meet those needs, which has major consequences for public policy~\citep{beard2016world}. 

Falls among older adults are a rising health risk globally, mainly because of accelerated population aging. Falls are the leading cause of injury-related hospitalizations and deaths among people aged 65 and older in many regions. For those 70 and older, fall-related deaths are the top category of injury-related fatalities~\citep{james2020global}. With an aging population and high fall rates among older adults, reducing the impact of falls is crucial. Older adults may experience acute conditions, such as fractures from falls or infections, requiring urgent care. This adds further strain to the healthcare system, already burdened by chronic diseases and other age-related health issues that may accompany longer population life expectancy. A study by~\citet{houry2016cdc} highlighted that falls among older adults often go undetected for extended periods, especially when individuals live alone. 

From the perspective of older adults, knowing that a fall will be detected quickly can help alleviate the fear of falling. This fear is common among older adults and can restrict their mobility and independence. Being assured of support in case of an accident can boost their confidence, encouraging a more active and mobile lifestyle, which benefits their overall health and well-being. From the perspective of professional caregivers and family members, near real-time alerts can facilitate better care management, ensuring that help is provided when it is needed most. Additionally, analyzing fall patterns can help identify risk factors and implement preventive measures, such as physical therapy, environmental modifications, or changes in medication. From the perspective of policymakers, preventing fall-related injuries and associated complications can help reduce healthcare costs. Acute care situations, such as emergency room visits, extended hospital stays, and rehabilitation, can be avoided by enabling older patients to stay at home safely for longer and avoiding costly long-term care facilities.
Delays in post-fall assistance increase the risk of severe injury or death. In fact, post-fall outcomes are critical. Research shows that if help does not arrive within 30 minutes, complications like dehydration and pressure sores are more likely. Prompt intervention is crucial~\citep{houry2016cdc,mikos2021falls}. Fall detection systems that quickly alert caregivers are essential for enhancing safety and supporting independent living in older adults. These tools are no longer optional; fall detection systems are a critical need.
% Falls are becoming a significant problem for older people worldwide.
% mainly because of rapid demographic aging and delayed fall recognition due to living alone. People aged 60 and above will represent more than 20 percent of the global population, with low and middle-income countries facing the steepest increases in aging-related care needs, by 2050~\cite{bhat2024prevalence}. Also, a study by Houry et al. \cite{houry2016cdc} highlighted that falls among older adults often go undetected for extended periods, especially when individuals live alone. 

Though related, fall prevention and fall detection serve different roles in older adult care. The primary goal of fall prevention is to reduce the risk of falling by identifying and modifying contributing factors. These factors include muscle weakness or poor balance, often addressed through exercise programs or environmental modifications~\citep{ajibade2025falls}. Fall detection systems, on the other hand, are designed to detect a fall once it occurs and trigger a rapid response, to address negative post-fall outcomes \citep{hardani2020autumn}. While fall prevention is proactive, its effectiveness can be limited due to unpredictable external factors and inconsistent adoption among elderly users, such as pets or leaving the walker behind.

Fall detection approaches can be summarized into wearable sensors, vision systems, and ambient sensing systems, each with strengths and limitations. Wearable sensors like accelerometers and gyroscopes are commonly utilized based on the ease of portability and the possibility of immediate data transmission, but their accuracy is heavily reliant upon the compliance of the user and regular device charging. Additionally, abrupt but inconsequential movements such as sitting or reclining may trigger false alarms \citep{delahoz2014survey,ramachandran2020survey}.

In this paper, we address a significant shortcoming of current fall detection systems: their absence of effective, multi-modal verification mechanisms that reduce false alarms without compromising user privacy or response latency. Most solutions today are based on a modality model, which hinders their capacity to differentiate genuine falls from routine activities and may lead to either alert fatigue or missed events.
Our work proposes a multi-stage framework that leverages wearable Inertial Measurement Unit (IMU) sensing, Received Signal Strength Indicator (RSSI)-based indoor location estimation, and visual verification assisted by a mobile robot. This multi-step pipeline allows the system to detect, localize, and verify the occurrence of falls more accurately, demanding visual confirmation only where needed to maintain user privacy and avoid spurious alarms.

In response to the challenges described above, our research makes the following key contributions: 

\begin{itemize}
    \item \textbf{We propose a novel privacy-preserving framework for fall detection} that integrates sensing, location, and robot vision to achieve high reliability, leveraging unlabeled data in a semi-supervised context and confirming fall detection via vision-based techniques.
    \item \textbf{We evaluate the framework across diverse experiments}, demonstrating its effectiveness in distinguishing falls from normal activities and its adaptability to the environment.
    \item \textbf{We develop an edge-cloud system} that provides prompt, effective notifications for real-world fall detection scenarios.
    \item \textbf{We advance a privacy-preserving strategy} that restricts invasive monitoring in high-risk situations.
\end{itemize}

These components collectively provide a scalable and feasible solution for the care of independent older individuals. The remainder of this paper is organized as follows: Section~\ref{sec:related_works} reviews the literature on federated learning, semi-supervised learning, vision-based fall detection, and robot-assisted fall detection frameworks. Sections~\ref{sec:method} and~\ref{sec:impl} detail the methodology and implementation of our fall detection framework. Section~\ref{sec:experiments} presents the experimental setup. In Section~\ref{sec:results}, we outline our experimental results, and Section~\ref{sec:discussion} presents a discussion of these results. Finally, Section~\ref{sec:conclusion} concludes by summarizing our contributions and outlining future work.
% \section{Motivation}
% \label{sec:motivation}

% Falls are the leading cause of injury-related hospitalizations and deaths among people aged 65 and older in many regions. For those 70 and older, fall-related deaths are the top category of injury-related fatalities~\cite{james2020global}. With an aging population and high fall rates among older adults, reducing the impact of falls is crucial.

% From the perspective of older adults, knowing that a fall will be detected quickly can help alleviate the fear of falling. This fear is common among older adults and can restrict their mobility and independence. Being assured of support in case of an accident can boost their confidence, encouraging a more active and mobile lifestyle, which benefits their overall health and well-being. From the perspective of professional caregivers and family members, near real-time alerts can facilitate better care management, ensuring that help is provided when it is needed most. Additionally, analyzing fall patterns can help identify risk factors and implement preventive measures, such as physical therapy, environmental modifications, or changes in medication. From the perspective of policymakers, preventing fall-related injuries and associated complications can help reduce healthcare costs. Acute care situations, such as emergency room visits, extended hospital stays, and rehabilitation, can be avoided by enabling older patients to stay at home safely for longer and avoiding costly long-term care facilities.
    
\section{Related Work}
\label{sec:related_works}
Research has identified that over 60\% of older individuals fail to participate consistently in prevention activities for falls due to cognitive, physical, or motivational restraints \citep{foo2025ai}. Additionally, even with preventive measures, 30\% to 40\% of individuals aged 65 or older continue to experience at least one or more falls per year, citing the inevitability of most falls \citep{sona2024iot}. Early detection is crucial in such instances. A delay of even 30 minutes in assistance can drastically raise the possibility of death or permanent disability \citep{elaoud2024explainable}. These challenges highlight the complexities in determining which fall detection frameworks are most practical and capable of delivering the best protective measures. The subsequent analysis explores the literature on federated learning, semi-supervised learning, vision-based fall detection, and robot-assisted fall detection frameworks, offering context for our research contributions.

\subsection{Federated Learning in Fall Detection}
Federated Learning (FL) is a distributed machine learning (ML) approach where multiple clients collaborate to train an ML model. In this setup, training data is decentralized to ensure data is kept private on individual devices or servers. It is used to train local models, which are updated/aggregated centrally, and redistributed to clients in the setup. With this approach, original localized data is never exchanged. In contrast, centralized learning approaches require a central raw data store from which to train ML models, increasing the vulnerability to breaches and privacy concerns~\citep{zhang2021survey}. Various studies have explored FL's potential in healthcare, particularly for fall detection and personalized monitoring.

FL, which has been applied to Human Activity Recognition (HAR), addresses privacy concerns by training models locally on user devices, thereby reducing communication costs and safeguarding personal data. \citet{sozinov2018human} explored different client data distributions and found FL yielded acceptable accuracy compared to centralized learning. However, complex models such as deep neural networks incur higher communication overhead in FL, making lower-complexity models more suitable for communication-sensitive applications. Thus, selecting a model requires balancing performance with deployment constraints.

\citet{wu2020mobile} proposed a two-level fall detection system. First, a lightweight detector on a smartwatch filtered out Activities for Daily Living (ADL) events using the root mean square of accelerometer data, sending potential fall signals to the cloud via Transmission Control Protocol (TCP). Second, an ensemble decision tree on the cloud accurately identified fall events. In this work, the approach was trained and tested primarily on publicly available datasets, with one practical dataset included to enhance evaluation. The Fall-detection Ensemble Decision Tree (FEDT) was the ensemble method proposed to identify falls. Experimental results demonstrated that the FEDT model improved sensitivity and specificity by 1-3\% compared to other ensemble learning models. Although the method did not use FL, given that the approach filtered out ADLs via a threshold, privacy was better preserved, as only potential fall events triggered data transmission, reducing unnecessary sharing of personal activity data.

\citet{yu_elderly_2022} proposed a fall detection method for older adults combining Federated Learning and Extreme Learning Machine (Fed-ELM). Their approach dynamically adjusts user-specific parameters and shares data securely, enhancing performance without compromising privacy. The method was first trained on the young subjects’ data from the SisFall dataset. Then, the model was enhanced by training on the incorrectly predicted samples from the older subjects. Fed-ELM achieves high accuracy, sensitivity, and specificity, outperforming traditional methods by leveraging federated aggregation and personalized updates. Experiments show robust, stable results across different users, effectively balancing detection performance and privacy.

\citet{wu2020fedhome} proposed FedHome, a cloud-edge FL framework for activity recognition that protects privacy by keeping health data local. It used a Generative Convolutional Autoencoder (GCAE) to handle data imbalance and non-independent and identically distributed (non-IID) issues, improving accuracy and reducing communication. By combining cloud and edge computing, FedHome enhances privacy, lowers latency, and supports personalized training. It outperformed both centralized and existing federated methods on balanced and imbalanced data.

\citet{ghosh2023feel} presented FEEL, a FL framework for activity monitoring, fall detection, and health recommendations without using video data. FEEL integrated the few-shot learning, user/context-based knowledge graphs, and clustering for personalization. Reinforcement Learning (RL) managed sparse data and diverse needs, while a custom wearable collected key health metrics. Experiments showed FEEL outperforms baselines, proving effective for smart hospitals and assisted living facilities.

Data collection and analysis at the edge can reduce latency, enhance privacy, and enable real-time decision-making without relying on cloud resources. 
\citet{sathya2023federated} proposed a FL-based fall detection system using data from a smart shoe and smartwatch. The dual-sensor approach improved accuracy and ensures privacy by sharing only model updates. Edge computing enabled real-time analysis and alerts, with fall confirmation through room camera images. Their method required continuous data transfer to an edge device for fall detection. However, this work was primarily conceptual, as it did not present any empirical results or experimental findings.

Data fusion is the process of integrating multiple sources of information to improve a model's predictive performance. It can play an important role in enhancing the accuracy of fall detection systems. \citet{qi2023fl} applied input-level data fusion by combining time-series data from wearable sensors with visual data from cameras. FL was employed to train the fall detection model. Experimental results on the UP-Fall dataset \citep{martinez2019up} highlighted the effectiveness of their approach, demonstrating that multimodal data fusion significantly outperformed single-modal methods in fall detection accuracy. Their framework achieved 99.927\% accuracy for binary classification of falls versus non-falls, and 89.769\% accuracy for multi-class fall activity recognition.

\subsection{Semi-Supervised Learning} 
Semi-Supervised Learning (SSL) utilizes a small amount of labeled data alongside a larger set of unlabeled data to enhance model performance, especially when labeled data is scarce or costly to obtain~\citep{van2020survey}. Fall detection systems often struggle due to their dependence on labeled user data, which is difficult to acquire in real-world settings. The labeling process can be intrusive, impractical, and poses privacy risks~\citep{ji2022sifall,diao2022semifl,tashakori2023semipfl,zhao2021semi,xiao2023towards}. Additionally, few existing approaches function effectively without labeled data while ensuring user privacy, underscoring the need for more efficient and privacy-preserving fall detection systems.

\citet{patricia2024semi} introduced a semi-supervised ensemble learning model for HAR by using distance-based clustering to identify distinct activity patterns, which were then classified using supervised techniques, showing favorable results, outperforming current methods. This approach offered interesting potential for advancing smart home technologies and healthcare-related applications and highlighted the value of hybrid and semi-supervised approaches.

\citet{yu2021fedhar} introduced FedHAR, a personalized federated HAR framework. The semi-supervised online learning approach is used to address privacy, label scarcity, real-time processing, and heterogeneity challenges. FedHAR uses a hierarchical attention model and a novel unsupervised gradient algorithm to boost recognition accuracy with minimal labeled and ample unlabeled data. An unsupervised gradient aggregation strategy is employed to address concept drift and convergence instability in online learning. The authors perform a series of experiments to demonstrate FedHAR's superiority over existing methods. Additionally, a personalized approach, PerFedHAR, is explored, which further improves performance by fine-tuning models for each client.

Semi-supervised FL offers a promising solution by combining FL with SSL to use both labeled and unlabeled data~\citep{zhao2021semi,tashakori2023semipfl}. \citet{zhao2021semi} showed that this integration enhances performance while maintaining privacy. \citet{tashakori2023semipfl} introduced SemiPFL, which improves adaptability in heterogeneous environments through the use of unlabeled data. Current methods have not specifically focused on wearable-based fall detection. To address this, we propose a semi-supervised FL approach for fall detection systems that enhances accuracy while preserving user privacy, bridging the gap between effective detection and data security.

% %not SSL - Commented out - RR
% \textcolor{blue}{The paper presents SmartFall, an Android application that utilizes accelerometer data from smartwatches to detect falls in real-time, ensuring privacy by processing data locally on the smartphone without the need for cloud communication. The study compares traditional machine learning models such as Support Vector Machine (SVM) and Naive Bayes with a Deep Learning model, demonstrating the superior accuracy and generalization capability of the Deep Learning model in fall detection across multiple datasets. The Deep Learning model is based on a recurrent neural network (RNN) architecture, specifically using gated recurrent units (GRU), which effectively capture temporal data patterns from raw accelerometer readings. This approach allows the model to discern subtle features that are not accessible through conventional feature extraction methods, enhancing the detection accuracy and reducing false positives compared to traditional models. SmartFall employs a three-layer IoT system architecture, facilitating data privacy and scalability. The system consists of a smartwatch for data collection, a smartphone for processing and prediction, and optional server storage for archival and further analysis. The application features an intuitive user interface designed for elderly users, providing alerts and GPS information to caregivers in case of detected falls~\cite{mauldin2018smartfall}.}

\subsection{Vision-Based Fall Detection}
Vision-based systems that use cameras to analyze human posture and motion have different challenges, such as privacy and occlusion. However, advances in algorithms, sensors, and real-time computing have improved vision-based fall detection systems. Specifically, low-cost radar systems, advanced ML, real-time edge computing, and personalized privacy strategies collectively improve motion capture, reduce environmental constraints, enhance accuracy, and support user safety and system acceptance, especially in home environments~\citep{inturi2025technical}.

\citet{gaya2024deep} presented a systematic review of computer vision-based HAR and fall detection, offering a comprehensive overview of current state-of-the-art methods. The review primarily examined three categories of vision data: RGB, depth, and infrared. Depth and IR data enhance privacy by omitting RGB footage but suffer from low resolution, limited interpretability, and scarce data. Infrared cameras can be costly in comparison to other sensors. Skeleton poses and sequences are a prevalent choice for data representation in the literature due to their reduced size, user anonymization, and interpretability while maintaining effective performance in fall detection and human activity recognition. With regards to data, many datasets lack older adult participants, have small sample sizes, and include irrelevant activity classes, making them less ideal for older adult-focused HAR and fall detection.

Vision systems provide high-resolution spatial and motion analysis with high detection accuracy in good conditions. Nevertheless, they are also the subject of criticism for invasion of privacy due to around-the-clock video recording, and accuracy is compromised with weak lighting or obstructed lines of sight \citep{wang2020elderly, gutierrez2021comprehensive}. Ambient sensors like radar, pressure mats, and WiFi sensors are less invasive and have no user interaction. However, they also have problems, most notably high setup costs, significant calibration, and periodic signal interference that affect accuracy \citep{tahir2022iot, karar2022survey}.

A vision-based fall detection system is presented in~\citet{maldonado2019fallen}. In this work, the authors use a mobile-patrol robot equipped with a 2D image-based algorithm, combining deep learning (DL) for person detection and a Support Vector Machine (SVM) for fall classification, achieving high precision and recall rates on a new dataset, the Fallen Person Dataset (FPDS). A mobile robot with a camera detects individuals and determines if they have fallen. The dataset helps distinguish between fallen and lying individuals. The robot can relocate if the detection is uncertain, but it lacks navigation abilities and cannot detect falls outside the camera's view.

\subsection{Robot-Assisted Fall Detection Frameworks}
\citet{do2018rish} presented the development of a robot-integrated smart home for older adult care, combining smart home technology, a service robot, and remote caregivers to support independent living. The system combined three capabilities: (1) to estimate user location, a filter-based approach is used to fuse IMU readings with passive infrared sensor data; (2) for activity recognition, audio and location data are fed into a Dynamic Bayesian Network; and (3) for fall response, given the fall detection feature, a robot navigated to the user's location using the localization system. Key limitations in this work included privacy concerns from the use of multiple microphones, scalability challenges posed by sensor-intensive localization, particularly in larger homes, and insufficient detail regarding navigation and rescue procedures.

\citet{elwaly2024new} presented a newly developed robot designed to address specific challenges faced by older adults, such as continuous indoor tracking and fall detection. This work described the robot's hardware and software components, emphasizing the use of a 2D LiDAR, IMU, and camera for environment mapping, localization, and fall detection. The fall detection was performed using the YOLOv7 algorithm, achieving an accuracy of 96\%. The robot featured a modular, customizable design to meet individual needs, using a two-wheel differential drive and a custom Robot Control Unit (RCU) for component control. Experiments showed effective indoor path planning via Hector simultaneous localization and mapping and the RRT*, with 93.8\% positioning accuracy and high-precision fall detection. For the best fall detection accuracy, the robot should be placed between 185 and 220 cm away from the person. A primary drawback of this work was the limited camera coverage, such that obstacles may reduce detection accuracy. Additionally, there was insufficient explanation of the role of navigation and robot movement in fall detection, and details on the rescue process and how the robot assisted the fallen individual are lacking.

A collaborative fall detection system was proposed using a wearable device and a companion robot, focusing on minimizing privacy intrusion while maximizing the robot's sensing range in~\citep{liang2021collaborative}. A threshold was used by the wearable device to detect a potential fall. If exceeded, camera data was collected and sent to the robotic platform. The wearable device used data from a camera, accelerometer, and microphone for initial fall detection, while the companion robot confirmed it through video analysis using Convolutional Neural Networks (CNN) and Long Short-Term Memory (LSTM) algorithms. Experimental results showed an overall accuracy of 84\% for the video-based fall detection algorithm, with specific activities like falling backward and walking achieving up to 94\% accuracy. However, the system exhibited notable limitations. It demonstrated low accuracy in fall detection, and the robotic platform functioned solely as a stationary edge device, lacking mobility and thus serving only as a computational resource within the environment.

\citet{capra2020assessing} explored the feasibility of augmenting traditional fall detection systems using Ultra-WideBand (UWB) technology and a home robot, aiming to minimize false alarms in fall detection among older adults. This work integrates a UWB sensor into a wearable device for localization, aided by additional UWB modules installed throughout the environment. The proposed system employed an ML algorithm for fall detection using accelerometer data, achieving a sensitivity of 99\% and specificity of 97.8\%. It also leveraged a UWB tracking system to locate the older adult and the robot in both Line-Of-Sight (LOS) and Non-Line-Of-Sight (NLOS) conditions. Although considerable effort was spent explaining the localization technique, this contribution lacked details on navigation and fall detection. Additionally, false alarms were overlooked, which might arise in real-world scenarios due to the detection model being trained on a limited dataset.

A fall detection system was introduced by \citet{sumiya2015mobile} that utilized a mobile robot equipped with a Kinect sensor, which had been widely used in research for its depth-sensing capabilities. Their system enabled tracking and monitoring of older individuals in real-time. This system addressed privacy concerns and the inconvenience of wearables by using a robot that followed users from a fixed distance, detected falls through skeletal analysis, and sent notifications to caregivers. The system reduced false positives by distinguishing falls from non-fall activities based on the duration the vertical position remains below a threshold. Experimental results showed up to 80\% fall detection accuracy with the mobile robot, outperforming fixed sensors. This approach relied on constant robot presence, which might cause discomfort for older adults.

\citet{kalinga2020fall} introduced a non-intrusive, vision-based fall detection system for older adults, using a service robot equipped with a Kinect sensor to monitor joint velocities and limb orientations, achieving 92.5\% accuracy and 95.45\% sensitivity. Similar to~\citet{sumiya2015mobile}, the solution depended on continuous robot presence, which might cause discomfort for older adults and impact battery capacity requirements. Additionally, it failed to address target loss or cases where the fall detection is confused by a user lying in bed or on the floor.

With regard to privacy, wearable devices are the most minimally invasive, while visual systems are the most challenging. Wearable systems can have high rates of false alarms due to everyday activities confusingly imitating falls throughout the day, while visual systems can provide high accuracy. Using each system alone compromises user confidence and the credibility of the response to emergencies, as users may distrust frequent false alarms or feel uncomfortable with constant visual monitoring. A holistic system incorporating a wearable device, FL, and robot-assisted visual event confirmation can optimize accuracy while enhancing overall reliability and enabling privacy preservation~\citep{xefteris2021performance, ni2024survey}.

\def\thickhline{\noalign{\hrule height1pt}}
\begin{table*}
    \centering
    \caption{An overview of fall detection frameworks with their datasets, learning methods, input types, and robot assistance.}
    
    \footnotesize
    \setlength{\tabcolsep}{3pt}
    \renewcommand{\arraystretch}{1.1} %
    \renewcommand\tabularxcolumn[1]{m{#1}}
    \begin{tabularx}{\textwidth}{>{\raggedright\arraybackslash}m{2.5cm}X|>{\raggedright\arraybackslash}m{1.5cm}|cc|cc|c}
    \thickhline
          &    &   & \multicolumn{2}{c|}{\textbf{Method}} & \multicolumn{2}{c|}{\textbf{Input}} &
          \\ 
          \textbf{Ref.} & \textbf{Contribution} & \textbf{Dataset} & \rotatebox{90}{FL} & \rotatebox{90}{SSL} & \rotatebox{90}{Vision} & \rotatebox{90}{Signal} & \rotatebox{90}{Robot-Assisted}  \\
    \hline
    \hline
        \citet{wu2020mobile}   & A two-level fall detection system using a smartwatch filter and cloud-based ensemble decision tree (FEDT), which preserves privacy by transmitting only potential fall events. & SisFall, MobiAct, Msys & & & & $\bullet$ & \\
    \hline
        \citet{yu_elderly_2022} & Fed-ELM, combining FL with extreme learning machines for elderly fall detection, training first on young subjects then adapting to older adults' incorrectly predicted samples. & SisFall & $\bullet$ & & & $\bullet$ & \\
    \hline 
        \citet{ghosh2023feel} & FEEL, a FL framework for activity monitoring and fall detection that integrates few-shot learning, knowledge graphs, and RL for personalization. & MobiAct & $\bullet$ & & & $\bullet$ & \\
    \hline
        \citet{sathya2023federated} & A conceptual FL framework for fall detection using dual sensors (smart shoe and smartwatch) with edge computing for real-time analysis, though without empirical validation. & TST-FB4FD, SmartFall & $\bullet$ & & & $\bullet$ & \\
    \hline
        \citet{qi2023fl} & A FL framework combining wearable sensor and camera data through input-level fusion, demonstrating multimodal approaches outperform single-modal methods. & UP-Fall & $\bullet$ & & & $\bullet$ & \\
    \hline 
        \citet{patricia2024semi}  & Semi-supervised ensemble learning model for HAR using distance-based clustering and supervised classification for smart home and healthcare applications. & CASAS Kyoto & $\bullet$ & & $\bullet$ & \\
    \hline
        \citet{yu2021fedhar} & FedHAR, a personalized FL framework for HAR using online SSL with hierarchical attention and unsupervised gradient aggregation.& RealWorld, HAR-UCI & & $\bullet$ & & $\bullet$ & \\ 
    \hline
        \citet{maldonado2019fallen} & A vision-based fall detection system using a mobile-patrol robot with combined DL person detection and SVM fall classification. &  FPDS & &  & $\bullet$ & & \\  
    \hline
        \citet{do2018rish} & Robot-integrated smart home system combining IMU-based localization, Dynamic Bayesian Network activity recognition, and autonomous robot navigation for fall response. & & &  & & $\bullet$  & $\bullet$ \\  
    \hline
        \citet{elwaly2024new} & A modular robot for elderly care using 2D LiDAR, IMU, and camera with YOLOv7 fall detection through Hector SLAM and RRT* path planning. & Custom & &  & $\bullet$ & $\bullet$  & $\bullet$ \\ 
    \hline
        \citet{liang2021collaborative} & Collaborative fall detection system using a wearable device for initial threshold-based detection and a stationary companion robot for CNN-LSTM video confirmation. & & &  & $\bullet$ & $\bullet$  & $\bullet$ \\ 
    \hline
        \citet{capra2020assessing} & Fall detection system augmented with UWB technology and home robot integration, using UWB sensors for localization and ML algorithms for accelerometer-based detection. & \citet{ozdemir2014detecting} & &  & & $\bullet$  & $\bullet$ \\ 
    \hline
        \citet{sumiya2015mobile} & A mobile robot fall detection system using Kinect skeletal analysis that follows users at a fixed distance. & & &  &  & $\bullet$  & $\bullet$ \\ 
    \hline
        \citet{kalinga2020fall} & Non-intrusive vision-based fall detection system using a service robot equipped with Kinect sensor to monitor joint velocities and limb orientations. & & &  &  & $\bullet$  & $\bullet$ \\ 
    \thickhline
    \end{tabularx}
    \label{tab:app-sum}
\end{table*}
\section{Methodology} \label{sec:method}
The following section provides a comprehensive overview of our fall detection framework, which builds upon our previous works \citep{azghadi2024adaptive,azghadisf2d}, beginning with a high-level description of the detection steps. Subsequently, we delve deeper into the individual components of the system. Finally, we categorize the system's processes into online and offline phases, emphasizing that certain steps must be completed before the system becomes operational.

\subsection{Overview}
\label{sec:method_overview}
\begin{figure}[ht]
      \centering
        \includegraphics[width=0.95\linewidth]{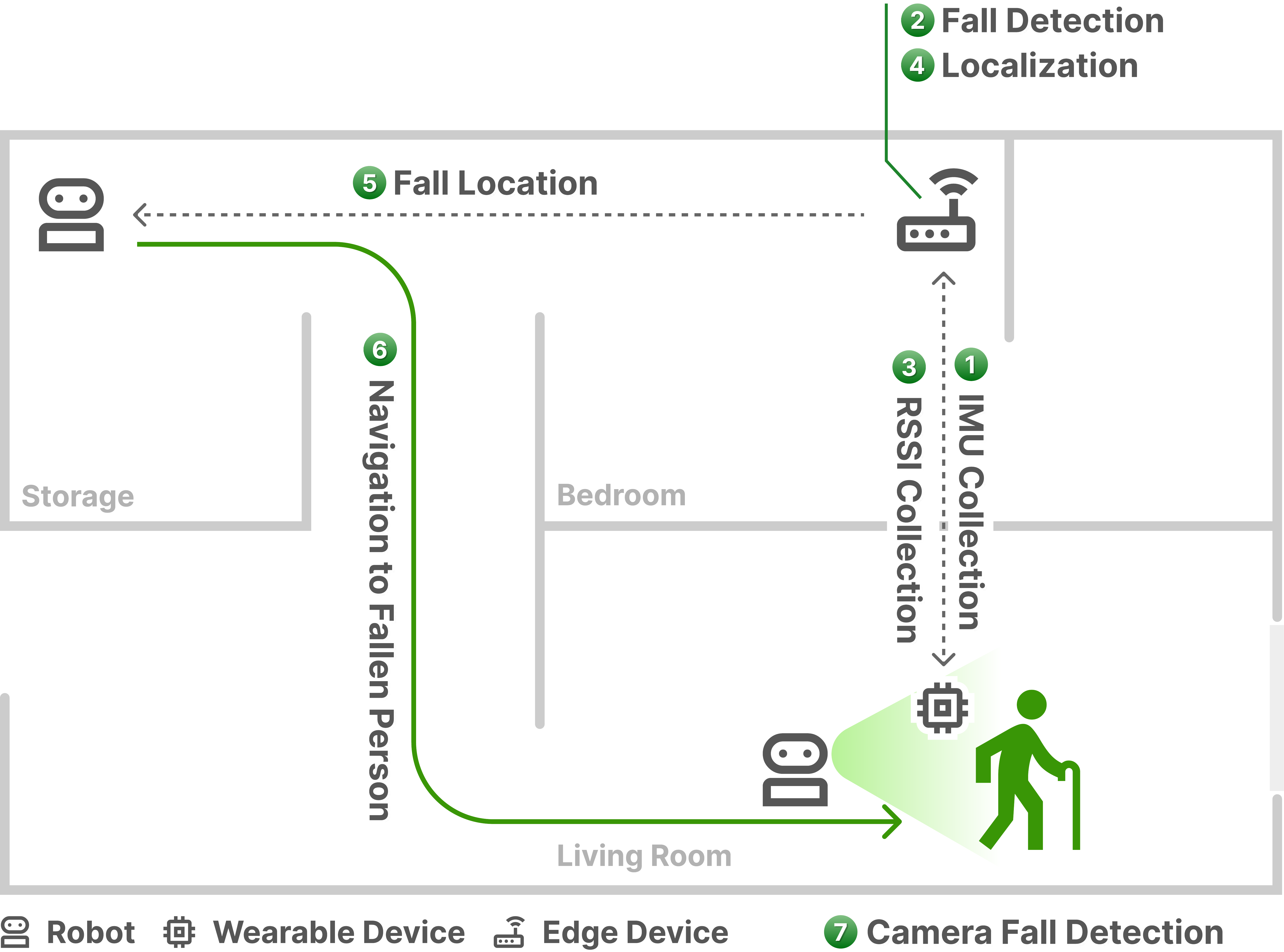}
      \caption{Our Fall Detection Framework Workflow}
      \label{fig:method}
\end{figure}

This section describes the fall detection framework during the online phase, in which all system components and processes are fully operational. Initially, as indicated by \circled{1} in Figure \ref{fig:method}, the wearable device captures and transmits IMU data corresponding to user movement to the edge device. The edge device then forwards this data stream to the fall detection model provided by the cloud server. If the model detects a potential fall event in the incoming data stream \circled{2}, the edge device requests that the wearable device collect and transmit RSSI information \circled{3}. Using this RSSI information along with a pre-trained localization model, the system estimates the user's precise location within the environment \circled{4}. At this stage, the edge device recognizes a potential fall event at a specific location. To confirm the fall and prevent false alarms, the edge device sends the estimated location coordinates to the robot \circled{5}. Utilizing a pre-generated map and the received coordinates, the robot navigates toward the user's position \circled{6}. During navigation and upon arrival, the robot actively searches for the fallen person by analyzing real-time camera streams through a detection model capable of distinguishing between sleeping and fallen individuals \circled{7}. Finally, if the robot confirms that the user has indeed fallen, the edge device contacts the cloud server to retrieve emergency contact information.

\subsection{Components of Framework} \label{sec:method_components}
\begin{figure}[t]
      \centering
        \includegraphics[width=0.95\linewidth]{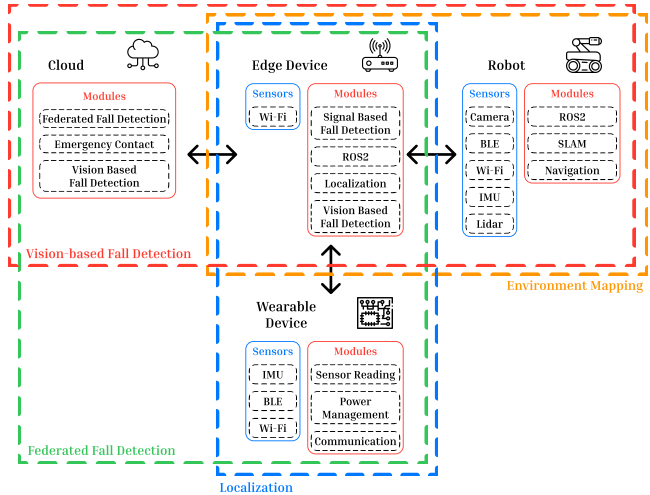}
      \caption{Fall Detection Components and Systems of the Proposed Framework and Their Connection}
      \label{fig:method_components}
\end{figure}

This section describes the various entities within our framework and their interactions. As illustrated in Figure \ref{fig:method_components}, our framework consists of four main components, each of which will be explained in detail. To highlight components involved in specific processes, we have used colored boxes around these components in Figure \ref{fig:method_components}. Furthermore, we categorize the processes in our system into two distinct subcategories: \textit{offline} and \textit{online}. Each component or process is assigned to one of these categories based on whether it operates before deployment (\textit{offline}) or during active system operation (\textit{online}).

Our fall detection framework consists of four main components. The first is a cloud server, which provides substantial computational resources and plays a critical role in training models using knowledge shared by individual users. The second component is an edge device located near the end-user, which offers sufficient computational capabilities for certain tasks and protects user privacy by eliminating the need to transfer data to the cloud. Third, we incorporate a wearable device that the user constantly carries; this device is essential for real-time applications such as fall detection and user localization. Finally, we employ an autonomous robotic platform, enhancing system reliability and flexibility by adding mobility to the system.

\begin{itemize}
    \item \textbf{Cloud Server:} Our cloud server has several important roles within the fall detection framework. Primarily, it is a crucial part of the semi-supervised federated fall detection system. As we will describe further in the process section, the cloud server aggregates autoencoder models trained by individual users and subsequently trains the final detection model. Additionally, the server is responsible for training the vision model used to identify fallen individuals. To reduce computational demands on users' edge devices, we train a global model using a benchmark dataset and then transfer this model to the edge devices. This approach enables robotic platforms to detect and confirm fall events accurately.
    \item \textbf{Edge Device:} This device is the most critical component of the framework, as it connects and coordinates all other parts to accomplish various tasks and applications. Additionally, it significantly reduces user privacy concerns by eliminating the need to transfer data outside the user's home. Each edge device trains an autoencoder model to capture different aspects of users' behaviours, which ultimately benefits all users. Apart from training models during the offline phase, the edge device also performs inference using machine learning models during the online phase. This enables the execution of multiple essential tasks, including the detection of fallen individuals, identifying fall events from wearable device data, and estimating the user's location.
    \item \textbf{Wearable Device:} The primary role of the wearable device is to collect data from the user and transmit it to the edge device. Although this task may appear straightforward, several challenges arise when implementing even a simple prototype. Issues such as connectivity, accuracy, power consumption, and packaging commonly complicate the design and development of wearable devices. To address these challenges, we developed our own wearable device specifically tailored to integrate seamlessly with the overall framework, ensuring it meets the diverse requirements of various processes.
    \item \textbf{Robot:} The final component of our fall detection framework is a robotic platform. We incorporated this robot into our system to address one of the primary challenges in the current fall detection framework: false positives. By leveraging the robotic platform, we can verify situations following positive detections from the wearable device and edge devices, effectively confirming whether a fall event has genuinely occurred. This approach allows us to label fall samples with greater confidence, significantly reduces false alarms, and prevents unnecessary emergency calls—all while ensuring minimal invasion of users' privacy.
\end{itemize}

\subsection{Processes of Framework}
\label{sec:method_processes}
The following section outlines the primary processes within our framework's systems. First, we describe the offline processes, beginning with the semi-supervised federated fall detection system. Next, we discuss the mapping and training of the localization models, a crucial step toward reducing false alarms. Additionally, we present a vision-based model designed specifically for detecting fallen individuals. Finally, we address the online processes, including real-time fall detection, localization, and fall event confirmation.

\subsubsection{Offline}
The processes in this category are designated as Offline because they occur prior to real-time fall detection in real-world scenarios. These processes involve training and constructing models that will later be deployed during the online phase.

\begin{itemize} 
    \item \textbf{Offline Semi-Supervised Federated Fall Detection:} \label{sec:method_processes_sf2d}
    One of the key tasks in our framework is detecting fall events using only wearable device data. To achieve this, we designed a semi-supervised federated fall detection approach. The process begins on the cloud server, where an autoencoder is initialized and then distributed to the users' edge devices. Each edge device independently trains this autoencoder using data collected from the older adults wearing the devices. At this stage, the training is entirely unsupervised. Each edge device trains the model for a few epochs and then transfers the model to the cloud for aggregation. In the cloud, we aggregate the partially trained models to create a new autoencoder and redistribute the model for a new round of training. After multiple rounds of training and aggregation, we obtained a federated autoencoder that incorporates knowledge learned from all participating users.

    In the next phase, we utilize a labelled benchmark dataset available in the cloud, which includes various types of falls and ADL, to train a fall detection classifier. The encoder part of the federated-trained autoencoder is frozen and employed as the backbone for this new detection model. After training the classifier, we distribute it to the users' edge devices, enabling online fall detection capability. This federated, semi-supervised approach ensures user privacy by eliminating the need to transfer users' private data to a centralized location for training.
    \item \textbf{RSSI Map Creation:}
    Localization and mapping are closely related yet distinct concepts. Depending on the device's capabilities and required performance, localization can be performed either with or without a map or simultaneously as we build the map. When localization and mapping occur concurrently, the process is known as Simultaneous Localization and Mapping (SLAM). The SLAM technique enables a robot or a system to build a map of an unknown environment while simultaneously determining its location within that map. However, this simultaneous approach is not always feasible because some devices, such as our wearable device, are incapable of generating a map or localizing themselves independently. Therefore, it is necessary to create a map beforehand, providing reference points that enable the wearable device to determine its position accurately.

    Our wearable device is equipped with Wi-Fi and Bluetooth Low Energy (BLE) modules, which can be used for localization. Although the device also includes an IMU module, relying solely on the IMU for localization can result in significant positional errors over time due to error accumulation. To address this limitation, we decided to use our robot to generate a map before launching the fall detection system. Using a SLAM algorithm, our robot creates a 2D map of the environment while simultaneously localizing itself. We then combine this spatial map with the BLE data collected by the robot to produce a refined map. This resulting map represents the signal strength from various BLE anchors at specific locations throughout the environment. Such a map is particularly beneficial for smaller devices that lack high-precision sensors, enabling them to localize effectively within their surroundings.

    % Fingerprinting?
    \item \textbf{Localization Model Training:}
    With a map containing signal strength information from different access points, we can build a localization model. The model's input will consist of a list of visible access points and their respective signal strengths, as these are the only relevant pieces of information that our wearable devices can reliably collect for localization. Therefore, before launching the fall detection system, we must construct a localization model based on the RSSI map we've previously generated.

    We designed a neural network capable of estimating the location of edge devices using this RSSI information. The output of our model is a pair of continuous coordinates, $x$ and $y$, representing the device’s estimated position on the spatial map previously generated by the robot. This location estimation can subsequently facilitate robot navigation, guiding the robot toward the user in the event of a fall.
    \item \textbf{Offline Vision-Based Fallen People Detection:}
    One of the crucial steps in our proposed framework is detecting fallen individuals using a camera mounted on the robot. Accomplishing this task requires training and developing a vision-based model capable of distinguishing between fallen people and those who are simply resting or sleeping, utilizing environmental cues such as the presence of mattresses or sofas.

    To achieve this, we trained the model using an existing dataset on the cloud and subsequently transferred the trained model to all edge devices. This approach primarily conserves energy at the user's end, as training such a model locally on each edge device would be both time-consuming and energy-intensive. Moreover, since all users require this functionality, deploying a pre-trained model ensures consistency and efficiency.

    Another critical consideration is the model's size, as it must be compact enough for efficient deployment on robots and edge devices to enable real-time detection.
\end{itemize}

\subsubsection{Online}
In this section, we describe the key processes involved in detecting and assisting fallen individuals during real-time system operation. These online processes utilize the models and data developed in the offline phase to ensure accurate and timely responses to fall events.

\begin{itemize}
    \item \textbf{Online Semi-Supervised Federated Fall Detection:} At this stage, a federated fall detection model has already been trained and deployed on the edge device. The wearable device continuously collects real-time IMU sensor data, batches it into half-second intervals, and transfers these data batches to the edge device. Upon receiving these batches, the edge device uses the fall detection model, preloaded into GPU memory, to determine whether the data indicates a fall or represents an activity of daily living.
    \item \textbf{Wearable Device Localization:}
    If the federated model detects a fall event, the edge device immediately sends a command to the wearable device to start collecting RSSI samples. These samples are then transmitted back to the edge device, where the previously trained localization model uses the RSSI data to estimate the precise location of the fall.
    \item \textbf{Robot Navigation:}
    Once the edge device estimates the fall location on the spatial map, it initiates autonomous navigation by either directly controlling the robot or transmitting the target location to the robot, allowing it to navigate by itself. Depending on the robot's available computational resources, the navigation task can be executed locally or offloaded to the edge device. In both cases, a cost map is created using the previously generated spatial map combined with live sensor data from the robot. The optimal path to the destination is determined based on this cost map. Given that the real-world environments (e.g., homes of elderly users) are dynamic, and the map may not be frequently updated, the robot must utilize all available sensor data alongside the stored map to effectively avoid obstacles and safely reach its destination.
    \item \textbf{Online Vision-Based Fallen People Detection:}
    \label{sec:method_processes_VFPD}
    As soon as the robot begins navigating toward the estimated fall location, it transmits a real-time camera stream back to the edge device. This video stream is processed by the vision-based fall detection model, previously trained in the cloud and now running on the edge device. The model continuously analyzes images as the robot navigates to detect the fallen individual. Two outcomes are possible:
    \begin{enumerate}
        \item If the detected individual has not fallen and the signal-based fall detection produces a false positive, the robot returns to its station. The recorded data is labelled as a false positive and can later be used to improve the fall detection model further.
        \item If the robot confirms that the person has indeed fallen, the edge device immediately sends an alert to the cloud server, triggering an emergency response by contacting either emergency services or the individual's relatives to request assistance.
    \end{enumerate}
    
\end{itemize}
\section{Implementation} \label{sec:impl}
In the following section, we will describe the implementation details of the mentioned processes in the \ref{sec:method_processes}.

\subsection{Semi-Supervised Federated Fall Detection}
\label{sec:impl_sf2d}

As described in Section~\ref{sec:method_processes_sf2d}, our federated fall detection framework begins by initializing an autoencoder network $\mathbf{A}$ with parameters denoted by $\theta_{\mathbf{A}}^{(0)}$ on the cloud server. The autoencoder consists of three stacked Long Short-Term Memory (LSTM) layers \cite{hochreiter1997long}. Once initialized, the global parameters $\theta_{\mathbf{A}}^{(0)}$ are distributed to each user's edge computing device $\mathbf{E}_{i}$, associated with user $\mathbf{U}_i$, where $i \in \{1, 2, \dots, N\}$.

Data collection involves capturing sensor measurements from wearable devices, which each user $\mathbf{U}_i$ wears. These devices integrate an accelerometer and a gyroscope, continuously recording acceleration $\mathbf{a}(t) = [a_x(t), a_y(t), a_z(t)]^\top \in \mathbb{R}^{3}$ and angular velocity $\boldsymbol{\omega}(t) = [\omega_x(t), \omega_y(t), \omega_z(t)]^\top \in \mathbb{R}^{3}$ at discrete timestamps $t$. These sensor measurements form the raw data vector $\mathbf{x}(t) = [\mathbf{a}(t)^\top, \boldsymbol{\omega}(t)^\top]^\top \in \mathbb{R}^{6}$. To minimize wireless transmission overhead, the wearable device aggregates data over a short interval of duration $k$ (e.g., $0.5\,\text{s}$), creating an aggregated packet:

\begin{equation}
\mathbf{X}(t_j) = [\mathbf{x}(t_{k}), \mathbf{x}(t_{k-1}), \dots, \mathbf{x}(t_{1})],
\end{equation}

where $k$ represents the number of observations per packet. Each aggregated packet $\mathbf{X}(t_j)$ is wirelessly transmitted to the corresponding edge device $\mathbf{E}_i$. During the training phase, each edge device $\mathbf{E}_i$ collects and stores $J_i$ such packets, forming the local training set:
\begin{equation}
\mathcal{S}_i = [\mathbf{X}(t_{J_i}), \mathbf{X}(t_{J_{i}-1}), \dots, \mathbf{X}(t_{1})].
\end{equation}

Before training the autoencoder locally, each edge device preprocesses its collected dataset. The preprocessing includes resampling the data at a fixed frequency $f$ (e.g., every $50\,\text{ms}$) to standardize measurements and reduce noise. Subsequently, an Exponentially Weighted Moving Average (EWMA) and Savitzky–Golay filters are applied to further reduce noise. Finally, the filtered data is segmented into fixed-length sequences of size $l$, preparing consistent inputs for the autoencoder training. The preprocessed data for user $\mathbf{U}_i$ is defined as:
\begin{equation}
\mathcal{D}_i = 
\begin{bmatrix}
x'_{11} & x'_{12} & \dots & x'_{1l} \\[3pt]
x'_{21} & x'_{22} & \dots & x'_{2l} \\[3pt]
\vdots & \vdots & \ddots & \vdots  \\[3pt]
x'_{r1} & x'_{r2} & \dots & x'_{rl}
\end{bmatrix},
\end{equation}
where $x'_{rl}$ denotes the preprocessed sensor value after filtering, and $r$ represents the total number of training samples, calculated by:
\begin{equation}
r = \frac{k \times J}{f \times l}.
\end{equation}

Each edge device $\mathbf{E}_i$ trains the autoencoder locally for a single epoch on its processed dataset $\mathcal{D}_i$, updating the local parameters $\theta_{\mathbf{A}}^{(i)}$ via gradient descent:
\begin{equation}
\theta_{\mathbf{A}}^{(i)} \leftarrow \theta_{\mathbf{A}}^{(i)} - \eta \nabla_{\theta_{\mathbf{A}}^{(i)}} \mathcal{L}_i(\theta_{\mathbf{A}}^{(i)}, \mathcal{D}_i),
\end{equation}
where $\eta$ is the learning rate and $\mathcal{L}_i$ denotes the loss function evaluated on local data $\mathcal{D}_i$.

After local updates, all edge devices send their updated parameters $\theta_{\mathbf{A}}^{(i)}$ back to the cloud server. The server aggregates these updates using the Federated Averaging (FedAvg) algorithm to generate a new global model:
\begin{equation}
\theta_{\mathbf{A}}^{(\text{global})} = \frac{\sum_{i=1}^{N} r_i \theta_{\mathbf{A}}^{(i)}}{\sum_{i=1}^{N} r_i},
\end{equation}
where $r_i$ is the number of data points used by edge device $\mathbf{E}_i$, serving as a weighting factor in the aggregation.

This iterative federated training process continues for a predetermined number of epochs, resulting in a robust global autoencoder capable of producing meaningful feature representations. Crucially, user privacy is maintained throughout this process, as raw sensor data never leaves the user's local edge device.

Following the autoencoder training phase, the framework transitions to the classifier training stage. Here, the encoder component $\mathbf{En}$ of the trained autoencoder $\mathbf{A}$ acts as a feature extractor and is integrated as the initial stage of a classifier $\mathbf{C}$. To create a complete classifier, a Fully Connected Neural Network (FCNN) is added atop the encoder $\mathbf{En}$ to predict activity labels.

The classifier $\mathbf{C}$ is trained on a labeled benchmark dataset $\mathcal{L} = \{(\mathbf{x}_k, y_k)\}_{k=1}^{K}$, where $\mathbf{x}_k \in \mathbb{R}^{6}$ is an input sample, $y_k$ is its corresponding activity label (fall or non-fall), and $K$ is the number of labeled examples. The preprocessing steps for classifier training mirror those used in autoencoder training, including resampling, EWM filtering, and sequence segmentation. Additionally, a momentum-based calculation is specifically applied to fall event samples to enhance the distinction between fall and non-fall activities, thereby improving classification accuracy. This step involves trimming excess data segments post-fall detection to capture more discriminative features.

During classifier training, the encoder parameters $\theta_{\mathbf{En}}$ remain frozen to prevent overfitting and bias towards the specific characteristics of the benchmark dataset. Only the parameters of the FCNN layers, denoted by $\theta_{\mathbf{C}}$, are updated. The encoder parameters remain fixed to preserve generalizability and ensure optimal performance when deployed with real-world user data.

\subsection{Creation of RSSI Maps and Localization Model} \label{sec:impl_map}

We developed a localization system composed of a robotic platform, an edge computing device, and multiple anchor points, designed to accurately localize a wearable device that is equipped solely with a BLE module. The localization process initiates with the construction of an environmental RSSI map. In our previous study \citep{azghadi2024adaptive}, we proposed an innovative method for capturing Wi-Fi RSSI data, leveraging a robotic platform equipped with SLAM capabilities. This SLAM-enabled robotic platform navigates the target environment while estimating its precise location within the generated spatial map. By synchronizing positional data obtained from the robot with Wi-Fi RSSI measurements acquired by the onboard Wi-Fi module, we generated an RSSI fingerprint dataset. Subsequently, a neural network was trained using the collected fingerprint dataset to predict locations based solely on RSSI input. Although the system achieved high localization accuracy when utilizing the same Wi-Fi module employed during map generation, replacing the Wi-Fi receiver significantly increased the localization error.

Initially, we intended to integrate a cost-effective Wi-Fi module (ESP32-C6) on the wearable device to facilitate localization. Despite experimenting with multiple strategies for performance enhancement, our findings consistently demonstrated that the ESP32-C6 module introduced significant noise into RSSI measurements, compromising the overall system reliability. This inherent noise was observed through experimental evaluation, where RSSI values from a stationary access point displayed notable fluctuations under consistent, line-of-sight conditions. Consequently, despite repeated attempts at data refinement and model optimization, satisfactory localization performance with this low-cost Wi-Fi module remained unattainable.

Alternatively, BLE-based localization presents a viable solution. Unlike Wi-Fi, publicly available BLE access points are typically not stationary, necessitating the deployment of dedicated anchor points throughout the environment. We successfully adapted our previous fingerprinting methodology to the BLE domain by integrating anchors into our environments. Importantly, by deploying identical BLE modules on both the robotic platform and the wearable device, our BLE-based localization system demonstrated improved accuracy compared to Wi-Fi-based localization under identical environmental conditions and hardware constraints.

The data collection workflow for training the localization model begins with developing and preparing the robotic platform. Specifically, we employed the Yahboom Rosmaster X3 robotic system \citep{yahboom}, augmenting its functionality through custom-developed Python software integrated within the Robotic Operating System (ROS2) \citep{ros2_2022} framework. The robot's control architecture comprised distinct software nodes, each providing dedicated interfaces for sensors and actuators. Additionally, we developed a web-based user interface to simplify robot control via an edge computing device. By executing our ROS2-based controller, sensor data could be monitored in real time through tools such as ROS2 and Rviz, which proved indispensable for evaluation and debugging purposes.

We incorporated the \texttt{SLAM\_Toolbox} library into our robotic platform's control architecture. This library integrates odometry data with Light Detection and Ranging (LiDAR) sensor measurements to simultaneously build environmental maps and robustly localize the robot. Odometry is estimating the robot's position and orientation by measuring the incremental movements of its wheels or other actuators over time. Furthermore, we implemented a specialized ROS2 node to interface with an ESP32-C6 device through serial communication, capturing BLE RSSI data and publishing these measurements as standardized ROS2 messages. A separate node was designed to subscribe to sensor-generated ROS2 topics, aggregating positional and RSSI data for storage in ROS2 bag files—a common structured format for robot sensor data. Each message recorded in the ROS2 bag included precise timestamps, facilitating subsequent synchronization of robot positions with RSSI measurements.

Following data collection, the stored ROS2 bag data underwent preprocessing to generate a structured fingerprinting dataset. Initially, sensor readings were extracted, inspected, and cleaned for anomalies and missing values. Data synchronization was then performed using Dynamic Time Warping (DTW) \citep{giorgino2009computing} to harmonize differing sampling frequencies and temporal discrepancies between sensors. The resulting synchronized dataset contained multiple rows, each representing a unique spatial position ($x$, $y$) on the two-dimensional map. Alongside spatial coordinates, each entry included RSSI measurements from all anchor points. Notably, some positions exhibited missing RSSI values (NaN) due to weak or absent signals from particular anchors. To address this, missing entries were systematically replaced with the minimum measurable RSSI value, ensuring dataset completeness and consistency. A representative example of the fingerprinting dataset structure is illustrated in Table \ref{table:fingerprinting}.

\begin{table}[t!]
\footnotesize
\centering
\caption{WiFi Fingerprinting Dataset Format}
\begin{tabular}{c c c c c c}
\toprule
Timestamp & $X\_Pos$ & $Y\_Pos$ & $MAC_1$ & $\cdots$ & $MAC_N$ \\
\midrule
1707935831.6001 & 0.0000 & 0.0000 & 66.0 & $\cdots$ & NaN \\
1707935832.5993 & 0.0026 & 0.0034 & 66.0 & $\cdots$ & NaN \\ 
$\cdots$ & $\cdots$ & $\cdots$ & $\cdots$ & $\cdots$ & $\cdots$ \\
1707935963.4922 & 3.4581 & 10.101 & 70.0 & $\cdots$ & 83.0 \\
\bottomrule
\end{tabular}
\vspace{-0.2cm}
\label{table:fingerprinting}
\end{table}

The subsequent stage involves training a localization model using the preprocessed fingerprinting dataset. This trained model is subsequently deployed to an edge computing device, enabling real-time localization of the wearable device. Specifically, we designed a neural network architecture that accepts the RSSI values obtained from all available anchor points as input and produces spatial coordinates ($x$, $y$) as output predictions. Ensuring consistent data ordering during the training and inference stages is crucial; thus, we maintain a mapping between anchor indices and their corresponding MAC addresses. Furthermore, we developed a Python-based application running directly on the edge device, which exposes a localization API accessible by other system components. The system is fully prepared for real-time operational use once the localization model is trained, validated, and deployed on the edge device.

\subsection{Vision-based Fallen People Detection}
\label{sec:impl_vfpd}

Other than the localization system and signal-based fall detection systems that we explained, we need one more system to be prepared before entering the online fall detection phase. As we explained, a robotic platform will inspect the environment upon detecting a fall with the wearable device. The robotic platform must be able to detect people and distinguish between fallen people and people in natural positions. Therefore, we used a YOLO (You Only Look Once)-based~\citep{redmon2016you} on a specific fallen people dataset for this task.

YOLO is a real-time object detection algorithm that identifies and localizes multiple objects within an image using a single, fast pass through a neural network. Built on a CNN architecture, YOLO performs detection and classification simultaneously in one unified step, making it highly efficient for time-sensitive tasks. Over time, YOLO has evolved from its initial version, YOLOv1, to the latest YOLO12, with each version bringing improvements in accuracy, speed, and architectural design to better handle the increasing demands of real-time applications. The models support both pretrained weights, commonly trained on large datasets like COCO for quick and effective deployment, and the flexibility to be fine-tuned or trained from scratch on custom datasets, making YOLO highly adaptable to specific use cases such as fall detection or human activity recognition.

We trained and evaluated our detection models using the publicly available FPDS \citep{maldonado-bascon_fallen_2019}, which contains 2062 labeled images of falls and non-fall activities across eight varied environments. The dataset's diversity in camera perspectives, body orientations, actor sizes, and lighting conditions made it particularly well-suited for testing the generalization capabilities of our fall detection algorithm.

We have done multiple experiments on different YOLO model versions to identify which one can achieve the best performance for our problem. As YOLO models would have made the inference on the edge device, it is important to consider the execution time. Another important factor in selecting the best architecture and version of YOLO is the detection accuracy. Therefore, we trained and evaluated all detection models from the three latest versions of the YOLO (v10, v11, v12) and selected the best version. We will provide more details in the experiment section.

In the fall detection scenario, upon detecting a fall event with Semi-Supervised Federated Fall Detection, we will send the robot to inspect the area. We developed a ROS2 node on the robot that captures the camera output and sends image messages on the \textit{camera} topic. We also developed another ROS2 node on the edge device that can capture these live images and use the trained model to identify if there is a fallen person in the image or not. The ROS2 node on the edge device annotates the images and publishes them again on the \textit{camera topic}. Finally, based on the explained scenarios in the section \ref{sec:method_processes_VFPD}, the edge device ROS2 node can send a message to the server.

\section{Experiments}
\label{sec:experiments}
This section presents the experimental setup, including the devices utilized and the results corresponding to the processes outlined in the preceding sections. We begin by elaborating on the four primary components of our system, as introduced in Section~\ref{sec:method_components}. Subsequently, we examine each individual process described in Section~\ref{sec:impl}, with a focus on their implementation and outcomes.

This section aims to experimentally validate the performance, integration, and real-world feasibility of our multi-stage fall detection framework. Through a series of modular and end-to-end experiments, we assess each component, highlighting its individual contributions and overall system reliability. The goal is to demonstrate that the proposed architecture not only achieves high accuracy but also meets the practical requirements for deployment in privacy-sensitive, real-world environments.

\subsection{Framework Components}
Here, we provide a comprehensive account of each system component, emphasizing both hardware specifications and implementation details. A thorough description of the components employed in our experiments is included, given their critical role in system performance. Such detail is essential for ensuring the reproducibility and transparency of our research findings.

\begin{figure}[t] 
    \centering
  \subfloat[\small{Wearable Device}\label{fig:wearable}]{\includegraphics[width=0.33\linewidth, height=0.33\linewidth]{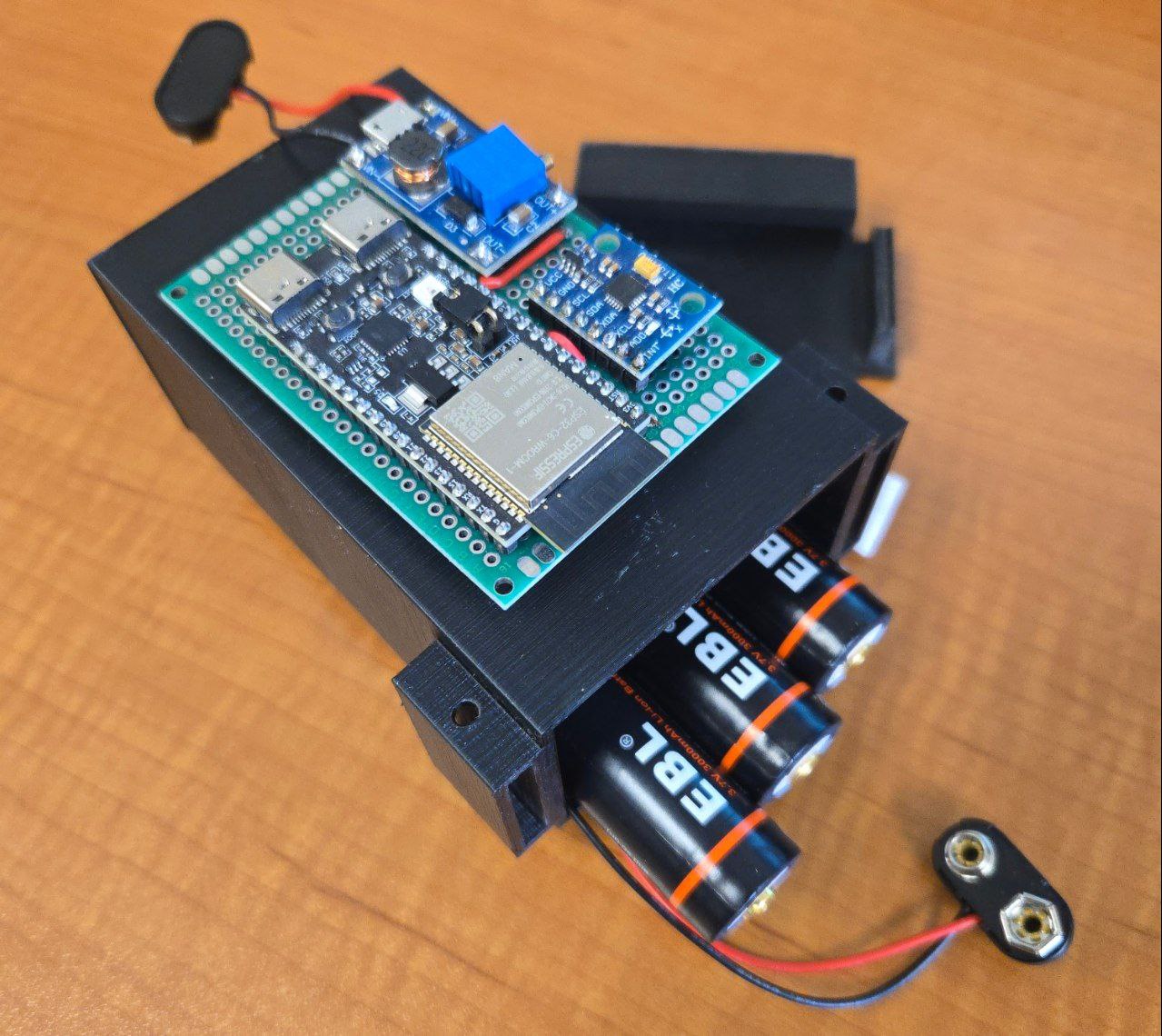}}
  \subfloat[\small{Edge Device}\label{fig:edge}]{\includegraphics[width=0.33\linewidth, height=0.33\linewidth]{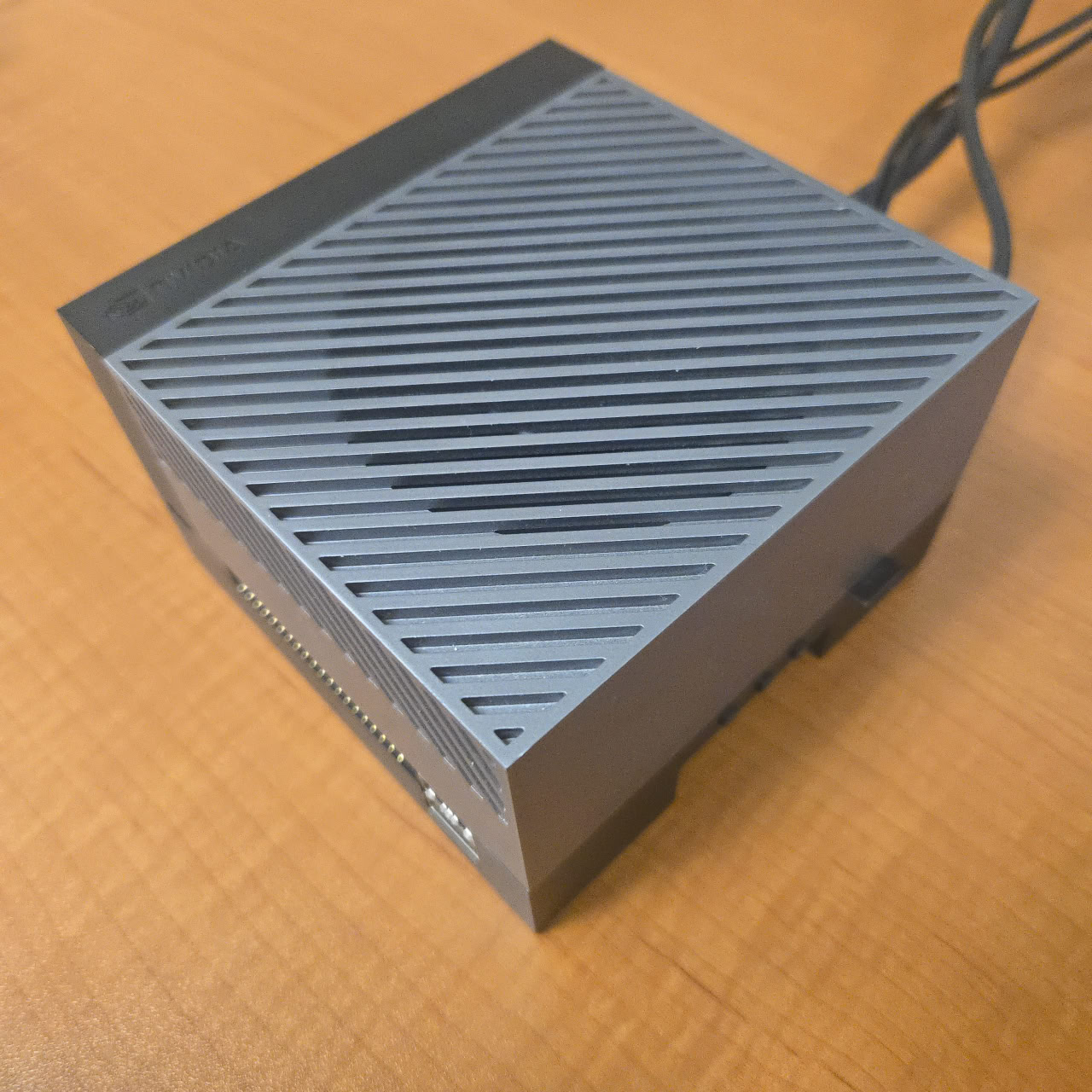}}
  \subfloat[\small{Robotic Platform}\label{fig:robot}]{\includegraphics[width=0.33\linewidth, height=0.33\linewidth]{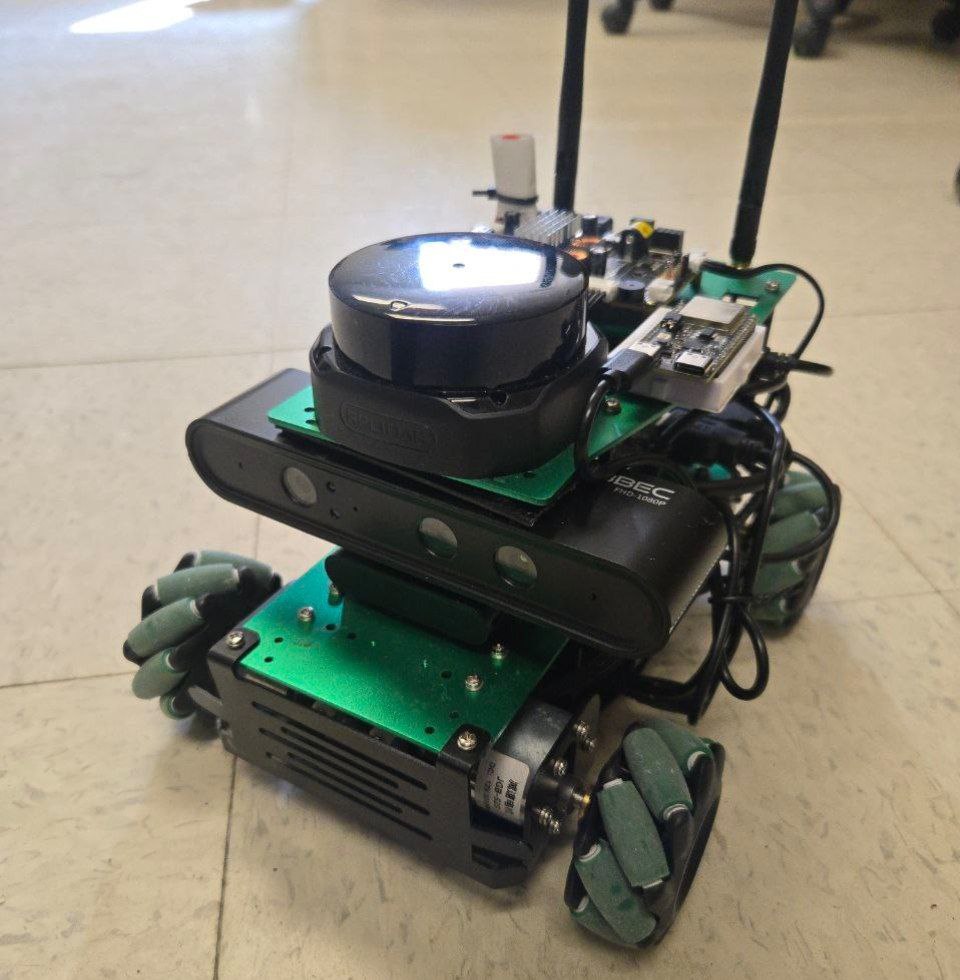}}        
  \caption{Components figure}
        \label{fig:experiements_components}
\end{figure}

\subsubsection{Cloud Server}
A high-performance computing system located in our office was used as the cloud server in our architecture. This machine is only connected to the university’s internal network and is not equipped with any wireless communication modules. The system is a \textit{Dell Precision 3660} workstation, equipped with a \textit{13th Gen Intel\textsuperscript{\textregistered} Core\textsuperscript{TM} i9-13900K} processor based on the \textit{x86\_64} architecture, featuring \textit{32 physical cores}. It includes \textit{128~GB} of RAM and \textit{2~TB} of storage. For GPU-accelerated tasks, the system is outfitted with an \textit{NVIDIA GeForce RTX 4090} graphics card with \textit{24~GB} of dedicated memory. This workstation was used to train all neural network models designated for cloud-side execution, as described in earlier sections.

\subsubsection{Edge Device}
The proposed architecture critically depends on the edge device, which serves as the central node facilitating connectivity among all system components. For this purpose, we employed the \textit{NVIDIA Jetson AGX Orin\textsuperscript{TM} 64GB Developer Kit}, intended to be deployed within the end-user's residence. This device is equipped with a \textit{12-core Arm\textsuperscript{\textregistered} Cortex\textsuperscript{\textregistered}-A78AE} processor based on the \textit{aarch64} architecture, comprising \textit{12 physical cores}. It features \textit{64~GB} of RAM and \textit{500~GB} of storage capacity. For GPU-accelerated computing tasks, the device integrates a \textit{2048-core NVIDIA Ampere architecture GPU} with \textit{64 Tensor Cores}, capable of delivering up to \textit{275~TOPS} (Tera Operations Per Second). The edge device plays a pivotal role in real-time data processing, local inference, local training, and communication with both cloud and wearable components.

\begin{table}[tb]
\footnotesize
\centering
\caption{Robotic Platform Sensors List}
\begin{tabular}{c c c}
\toprule
\textbf{Sensor} & \textbf{Model} & \textbf{Specification} \\ 
\midrule
2D LiDAR & \begin{tabular}{@{}c@{}} Slamtec \\ RPLIDAR S2 \end{tabular}    
         & \begin{tabular}{@{}c@{}} 360° coverage, 10~Hz scanning \\ Range: 0.05--50~m \end{tabular} \\
Wi-Fi   & Intel 8265 AC & Dual-band (2.4~GHz, 5~GHz) \\ 
IMU     & MPU-9250      & \begin{tabular}{@{}c@{}} Nine-axis sensor \\ Sampling rate: 400~Hz \end{tabular} \\
Camera  & \begin{tabular}{@{}c@{}} ASTRA \\ Pro Plus \end{tabular}  
         & \begin{tabular}{@{}c@{}} RGB-D camera, 30~fps \\ Depth range: 0.6--8~m \end{tabular} \\
Wheel Encoders & Hall-effect Sensors & Dual-sensor configuration \\
\bottomrule
\end{tabular}
\label{tab:robot_info}
\end{table}

\subsubsection{Wearable Device}
A custom wearable device was designed and implemented as part of the system architecture. The prototype comprises an \textit{ESP32-C6} development board, a motion tracking sensor (\textit{MPU6050}), and a step-up voltage converter (\textit{MT3608}). The \textit{ESP32-C6} functions as the central processing unit, communicating with the \textit{MPU6050} through the I2C interface to acquire motion data and leverage its integrated Wi-Fi module to transmit the data to the edge device. To ensure stable electrical connections, all components were mounted onto a protoboard. Additionally, we designed a 3D-printed box using \textit{PLA (Polylactic Acid)} filament to serve as housing for the wearable device. The device is powered by three \textit{ELB 18J Li-ion} batteries, each with a capacity of \textit{3000~mAh}. These batteries are connected in parallel to extend the operational lifetime of the wearable. The device was programmed using the \textit{Arduino} development environment, through which custom firmware was developed to manage sensor data acquisition and wireless communication.

\subsubsection{Robot}
We introduced a robot into the fall detection platform to provide real-time assistance and prevent false alarms. Our robotic platform is based on \textit{Yahboom ROSMASTER X3}~\citep{yahboom} as the primary physical system for data collection and localization tasks. This platform provides a robust hardware foundation with integrated support for sensors, actuators, and computational modules. Our robotic platform is outfitted with multiple sensing modalities, including a 2D LiDAR, an RGB-D camera, a motion tracking sensor, wheel encoders, and a Wi-Fi module. A detailed summary of the sensors and their specifications is provided in Table~\ref{tab:robot_info}. At the core of the platform is an \textit{NVIDIA Jetson Orin Nano}, which serves as the primary processing unit. This module interfaces with two auxiliary boards: (1) a robotic expansion board based on an STM32 microcontroller that manages low-level sensor input and wheel actuation; and (2) a USB hub that supplies adequate power and connectivity for peripheral sensors. The Jetson Orin Nano operates on \textit{Ubuntu 20.04} and runs the \textit{ROS~2 Galactic} distribution for middleware communication. As shown in Figure~\ref{fig:robot}, the robot employs omnidirectional \textit{Mecanum wheels}, enabling holonomic motion in both the $x$ and $y$ axes, as well as rotation about the $z$-axis. While this configuration offers enhanced maneuverability, it introduces additional complexity in modeling kinematics and estimating odometry, particularly when relying on wheel encoder data.

\subsection{Semi-Supervised Federated Fall Detection}
This section presents our experimental work for Semi-Supervised Federated Fall Detection. We begin by detailing the benchmark dataset utilized, along with the procedures for data cleaning and preprocessing. Subsequently, we describe the simulation-based training and report the corresponding results. Finally, we outline the design and implementation of a real-world clinical study conducted using our custom-developed wearable device.

\subsubsection{Dataset}
In order to evaluate our Semi-Supervised Federated Fall Detection, we utilized the SiSFall dataset \citep{sucerquia2017sisfall}, which offers a wide variety of both fall-related and daily living (non-fall) activities. Notably, it includes data collected from both young and elderly participants, making it suitable for evaluating age-diverse scenarios. The dataset comprises recordings of 15 types of falls and 19 types of ADLs, performed by 38 individuals across multiple trials (ranging from 1 to 5) with durations between 12 and 100 seconds, as summarized in \ref{tab:sisfall_labels}. 

Data were captured using three sensors—two accelerometers and one gyroscope—embedded in a wearable belt device positioned at the waist. All recordings were sampled at 200 Hz. For our analysis, we focused on data from one accelerometer and one gyroscope to perform the fall detection.

To ensure data quality and consistency across samples, a systematic cleaning pipeline was applied to the SisFall dataset to transform continuous sensor signals into structured, labeled segments suitable for machine learning models. After extracting raw motion data from the SisFall dataset, we converted raw sensor values using scaling factors derived from sensor resolution and range. Then we resampled the original data to a lower frequency, $50$ Hz, to reduce the computational cost. 

To reduce sensor noise and enhance signal quality prior to segmentation, two smoothing techniques were considered: 
\begin{itemize}
    \item \textbf{Exponential Weighted Moving Average} \citep{hunter1986exponentially}: This method applies exponentially decreasing weights to older observations, enabling responsive yet smoothed readings over time.
    \item \textbf{Savitzky–Golay Filtering} \citep{savitzky1964smoothing}: A second-order polynomial was fit over a sliding window of five samples, allowing for noise reduction while maintaining the integrity of critical features such as peaks or sudden changes, which are relevant for fall detection.
\end{itemize}
 
% EWMA reduces random sensor noise while staying responsive to sudden motion changes, making it ideal for detecting fall dynamics in real time. Savitzky–Golay filtering smooths the signal while preserving critical features like peaks and transitions, which are essential for accurately identifying fall events. 
These filters were explored experimentally and can be toggled based on the noise level of the raw data. Then, segmented each time-series signal into overlapping sliding windows of $40$ samples with a step size of $10$, corresponding to a $2$ seconds window. Each window was treated as an independent input sample for classification.

To support binary classification, all fall-related activities were labeled as class 1, and non-fall activities (ADLs) were labeled as class 0. We also applied normalization to scale sensor values into a standardized range of $[-1, 1]$ using min-max normalization. To reduce user bias in training, we adopted a \textbf{subject-independent} evaluation protocol, splitting users randomly into $70\%$ training and $30\%$ testing sets. This prevents data leakage and better evaluates generalization to unseen individuals.

\subsubsection{Benchmark Simulation}
We have done an experiment on the cleaned and processed dataset to evaluate our Semi-Supervised Federated Fall Detection on the benchmark dataset. Our experiment contains two parts. First, we trained and evaluated the system that we explained in the section \ref{sec:method_processes_sf2d} in a centralized way without using a federated learning algorithm, which we will name $\mathbf{CL}$. Second, we trained the autoencoder part with federated learning, and the classifier is trained in the cloud, which is called $\mathbf{FL}$. Notably, both parts are using the same data split to provide a consistent result. 

For both parts, we used $30\%$ of the dataset users as the labeled benchmark dataset in the cloud, known as $\mathcal{L}$. The rest of the dataset, $70\%$, is used for training and evaluating the autoencoder known as $\mathcal{D}$. Then we split the training users $\mathcal{D}$, into train and test with an $85\%$ to $15\%$ ratio known as $\mathcal{D}_{train}$ and $\mathcal{D}_{test}$.

In the $\mathbf{CL}$ scenario, we trained the autoencoder with the $\mathcal{D}_{train}$ dataset for $50$ epochs with \textit{Mean Squared Error} loss function to reduce the construction loss between the input and output signal. After training the autoencoder, the encoder part is frozen and fully connected layers are attached to it to create the classifier. For training the classifier, we used the other part of the dataset that has not been shown to the autoencoder, which is $\mathcal{L}$. We trained the classifier for $50$ epochs and used \textit{Categorical Crossentropy Loss} as the loss function. Finally, we can evaluate the classifier with the $\mathcal{D}$, as our goal was to train a classifier model that can distinguish the fall of users without requiring labeled data from that user.

On the other hand, in the $\mathbf{FL}$ scenario, the autoencoder training part is different. We split the $\mathcal{D}_{train}$ dataset based on the number of users in it, $N$, and trained the autoencoder assuming that there are $N$ Edge devices and each edge device has access to one user's data. We trained the autoencoder for $50$ rounds, and in each round, the model is trained for one epoch on each edge device. After training the federated autoencoder, the encoder part is frozen, and fully connected layers are attached to it to create the classifier similarly. After training the classifier, the trained model is again tested with the users' data $\mathcal{D}$. We used the Flower federated framework \citep{beutel2020flower} for this simulation.

\begin{table}
\centering
\scriptsize
\caption{SisFall Dataset Labels and Activity Descriptions}
\renewcommand{\arraystretch}{1.1}
\begin{tabular}{|c|p{5cm}|c|c|}
\hline
\textbf{Codes} & \textbf{Activities} & \textbf{Trials} & \textbf{Duration} \\
\hline
\multicolumn{4}{|c|}{\textbf{Fall Activities}} \\
\hline
F01 & Fall forward while walking caused by a slip & 5 & 15s \\
F02 & Fall backward while walking caused by a slip & 5 & 15s \\
F03 & Lateral fall while walking caused by a slip & 5 & 15s \\
F04 & Fall forward while walking caused by a trip & 5 & 15s \\
F05 & Fall forward while jogging caused by a trip & 5 & 15s \\
F06 & Vertical fall while walking caused by fainting & 5 & 15s \\
F07 & Fall while walking, using hands on a table to dampen fall, caused by fainting & 5 & 15s \\
F08 & Fall forward when trying to get up & 5 & 15s \\
F09 & Lateral fall when trying to get up & 5 & 15s \\
F10 & Fall forward when trying to sit down & 5 & 15s \\
F11 & Fall backward when trying to sit down & 5 & 15s \\
F12 & Lateral fall when trying to sit down & 5 & 15s \\
F13 & Fall forward while sitting, caused by fainting or falling asleep & 5 & 15s \\
F14 & Fall backward while sitting, caused by fainting or falling asleep & 5 & 15s \\
F15 & Lateral fall while sitting, caused by fainting or falling asleep & 5 & 15s \\
\hline
\multicolumn{4}{|c|}{\textbf{Activities of Daily Living (ADLs)}} \\
\hline
D01 & Walking slowly & 1 & 100s \\
D02 & Walking quickly & 1 & 100s \\
D03 & Jogging slowly & 1 & 100s \\
D04 & Jogging quickly & 1 & 100s \\
D05 & Walking upstairs and downstairs slowly & 5 & 25s \\
D06 & Walking upstairs and downstairs quickly & 5 & 25s \\
D07 & Slowly sit in a half-height chair, wait a moment, and get up slowly & 5 & 12s \\
D08 & Quickly sit in a half-height chair, wait a moment, and get up quickly & 5 & 12s \\
D09 & Slowly sit in a low-height chair, wait a moment, and get up slowly & 5 & 12s \\
D10 & Quickly sit in a low-height chair, wait a moment, and get up quickly & 5 & 12s \\
D11 & Sitting, trying to get up, and collapse into a chair & 5 & 12s \\
D12 & Sitting, lying slowly, wait, and sit again & 5 & 12s \\
D13 & Sitting, lying quickly, wait, and sit again & 5 & 12s \\
D14 & Lying on back, turn to side, wait, then return & 5 & 12s \\
D15 & Standing, slowly bending at knees, and standing up & 5 & 12s \\
D16 & Standing, slowly bending without knees, and standing up & 5 & 12s \\
D17 & Standing, get into a car, sit, and exit the car & 5 & 25s \\
D18 & Stumble while walking & 5 & 12s \\
D19 & Gently jump without falling (e.g., reaching high) & 5 & 12s \\
\hline
\end{tabular}
\label{tab:sisfall_labels}
\end{table}
\subsection{Creation of RSSI Maps and Localization Model}
This section presents the experimental methodology employed for generating RSSI maps and developing the indoor localization model. First, we describe the data acquisition process conducted within our office environment. Second, we outline the construction of the fingerprinting dataset from the collected raw RSSI values. Finally, we detail the training of the localization model and the implementation of the real-time localization procedure.

\subsubsection{Data Collection}

\begin{figure}[htb] 
    \centering
      \subfloat[\small{RViz application}\label{fig:rviz_crop}]{\includegraphics[width=0.95\linewidth]{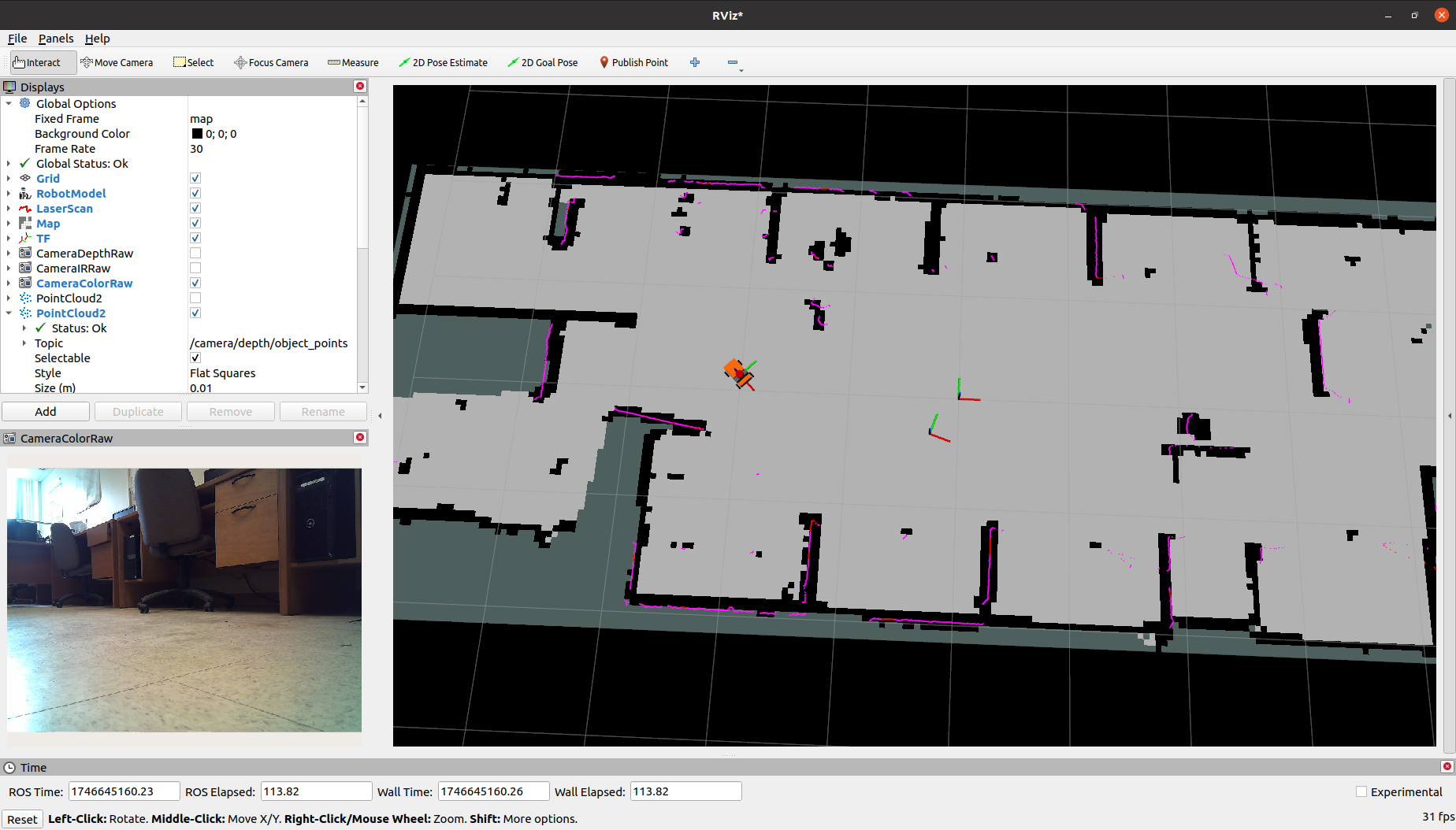}}
      \newline
      \subfloat[\small{Map and Odometry transform}\label{fig:rviz_zoom}]{\includegraphics[width=0.95\linewidth]{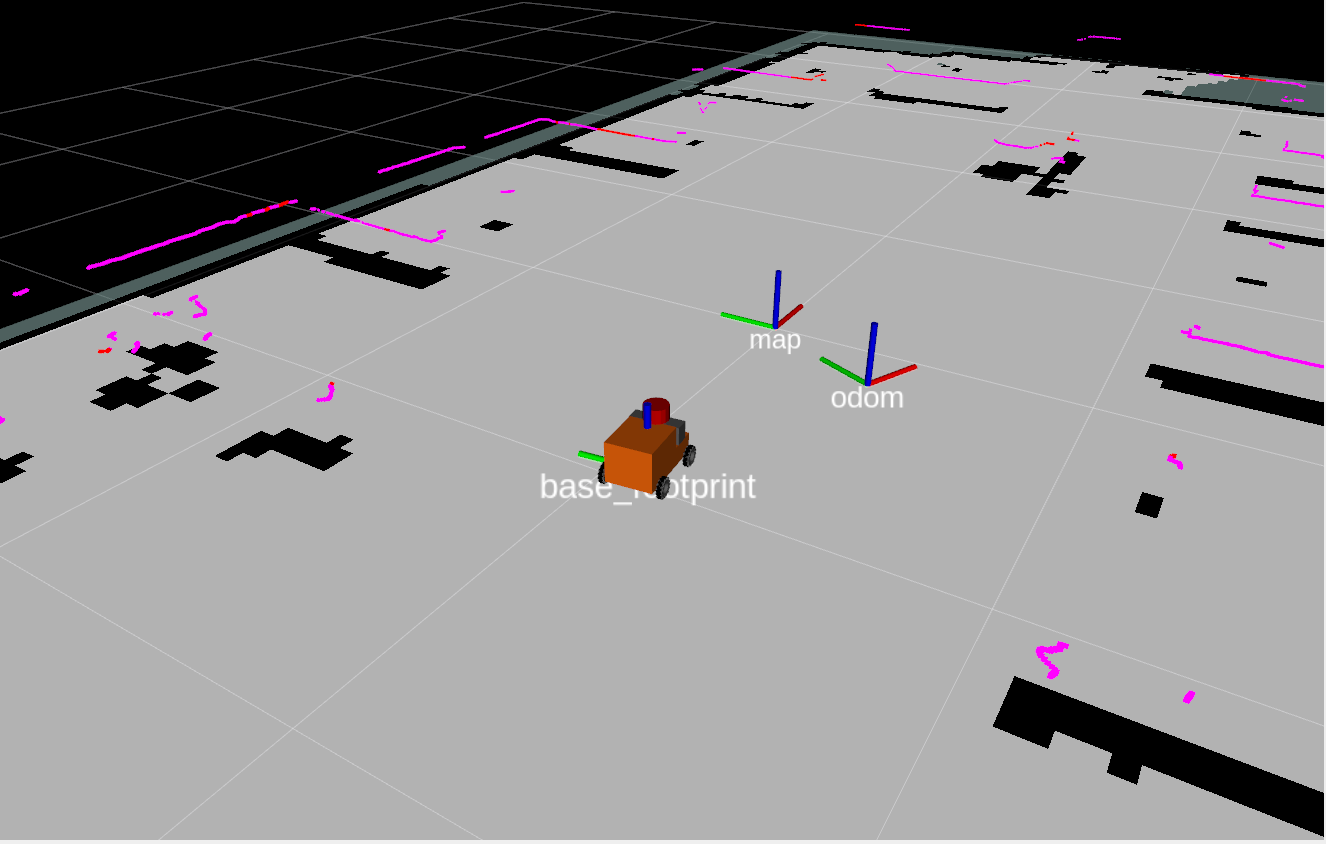}}
      \caption{RViz application during the data collection process}
    \label{fig:rviz}
\end{figure}

Data collection was conducted within our research lab at the university to evaluate the performance of RSSI map generation and the localization model. Multiple anchors were installed in the room, each programmed to emit BLE beacons at $100ms$ intervals. Five ESP32-C6 devices were programmed using the Arduino IDE to broadcast BLE beacons at the specified frequency. An additional ESP32-C6 device was configured to continuously receive these beacons and was installed on the robot to transmit beacon signal strength values to the robot via a serial connection.

In parallel, we developed a Python-based ROS2 node to interface with the ESP32-C6 receiver onboard the robot. This node listens to the serial port, parses the incoming BLE data, and converts it into standardized ROS2 messages. These messages are recorded in a ROS2 bag file alongside data from other onboard sensors, enabling synchronized data analysis. This setup allows for the collection of BLE beacon signals from multiple anchors, each tagged with the robot’s position at the moment of reception.

After initializing the anchors and configuring the robot, data collection was carried out by manually moving the robot to various locations within the room with the joystick. During this process, the system continuously recorded BLE beacons and stored them in a ROS2 bag file. The entire data collection process was monitored in real-time using RViz on an edge device connected to the same local network. Figure~\ref{fig:rviz} illustrates the RViz visualization of sensor outputs and the SLAM-generated map. Upon completion, both the ROS2 bag file and the map file produced by the SLAM\_Toolbox were transferred to the edge device for subsequent processing and analysis.

\subsubsection{Fingerprinting Dataset Creation}

The next step involves generating the fingerprinting dataset, beginning with the cleaning and preprocessing of the ROS2 bag file. In this file, all messages are stored with precise timestamps indicating the exact time of reception. Since different message types are logged at varying frequencies, temporal alignment was required to ensure consistency across data streams. To address this, we employed DTW to synchronize the data.

The synchronization process begins by applying DTW to the timestamp sequences of two message sets in order to determine the optimal temporal correspondence between samples. Once the alignment is computed, each sample in the set with fewer entries is matched to its closest counterpart in the larger set. This results in two aligned datasets of equal length, with each entry associated with a common timestamp, enabling accurate data fusion for localization.

Three types of messages were selected for synchronization using DTW to construct a precise fingerprinting dataset. First, we aligned the \textit{robot’s location} from the odometry node with the \textit{robot's drift} with respect to the map. The mentioned robot drift is based on the odometry node's accumulated error over time, which is calculated and published from the SLAM node. These two data streams were aligned to create a consistent, accurate location of the robot. Once corrected positions were obtained, we aligned them with the RSSI samples collected from the BLE receiver. As a result, each RSSI sample was associated with a corresponding corrected robot location, enabling the dataset to be easily structured in tabular form, as shown in Table~\ref{table:fingerprinting}.

\begin{figure}[t] 
    \centering
        \includegraphics[width=0.95\linewidth]{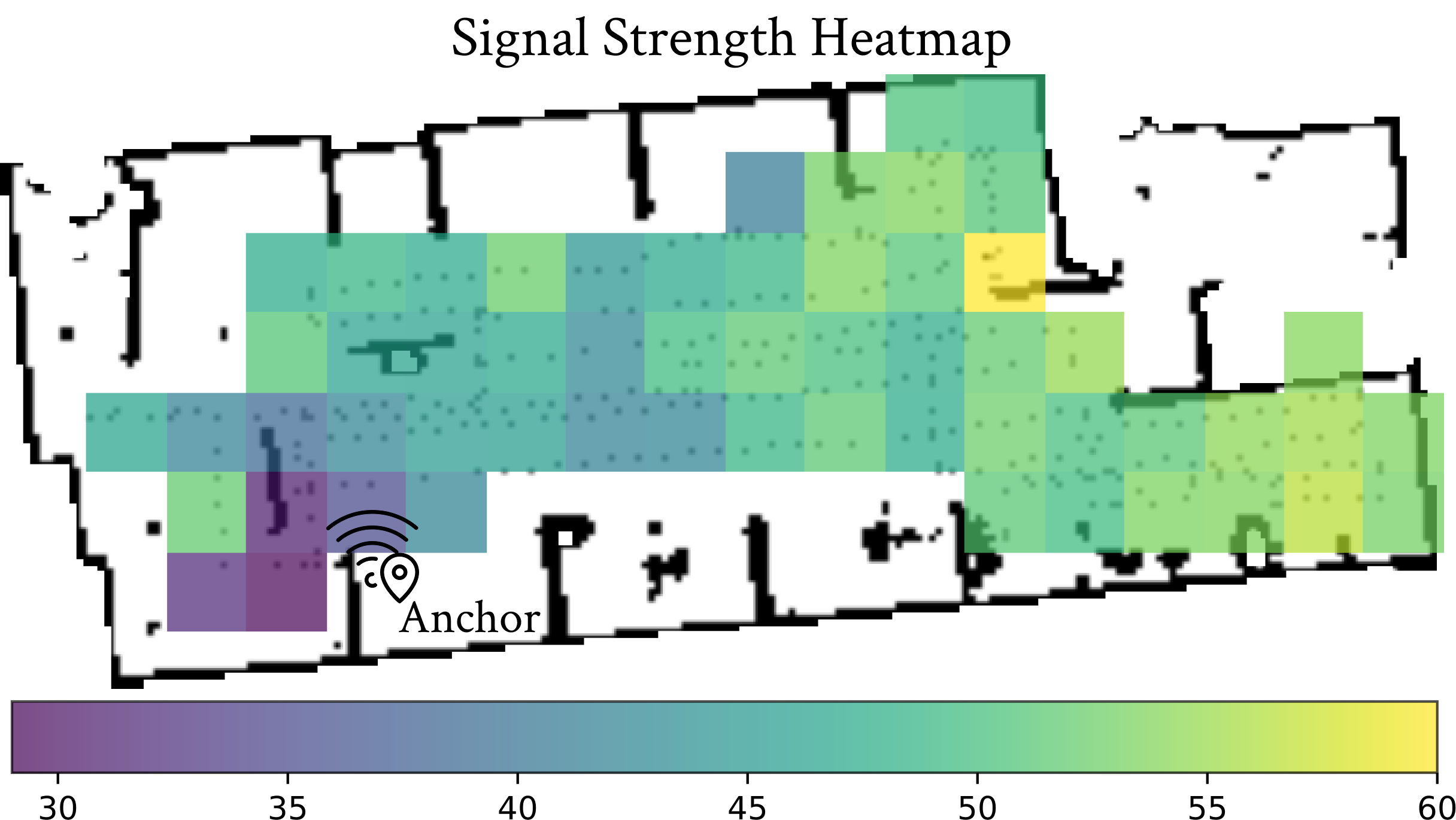}
        \caption{Overlapping the signal strength from one anchor with the sample collection location on the 2d generated map with lidar and SLAM}
        \label{fig:rssi_heatmap}
\end{figure}

To visualize the collected data, we generated a heatmap of the received signal strength from each anchor. This heatmap was overlaid on the geolocation map produced by the SLAM algorithm to illustrate the impact of obstacles and distance on signal propagation. Since the fingerprinting dataset is sparse and lacks measurements for every pixel on the map, we applied spatial averaging to improve coverage. Specifically, the map was divided into 8×8 pixel blocks, and the average RSSI value within each block was computed to construct the heatmap. This smoothing technique helped reduce noise and highlight signal trends across the environment. An example heatmap for one of the anchors is presented in Figure~\ref{fig:rssi_heatmap}.

\subsubsection{Localization Model}

Using the constructed fingerprinting dataset, we developed and trained localization models that estimate a device position based on the RSSI from multiple anchors. The objective of the localization model is to infer spatial coordinates from new RSSI inputs, enabling real-time indoor positioning. To facilitate both training and inference, we implemented a web-based Python application on the edge device. This application exposes APIs for training the localization model using the fingerprinting dataset and for estimating the device's position based on newly received RSSI samples.

The application incorporates a data processing pipeline that cleans the input dataset before training the model. In the initial stage, missing RSSI values—caused by anchors being out of range in certain areas—are replaced with a default value of $-100 dB$, representing the minimum detectable signal strength. Subsequently, the system generates a consistent mapping between the MAC addresses of the anchors and the input features of the model. This step ensures that the RSSI values are arranged in a uniform order, thereby preventing inconsistencies in model input during both training and inference. Finally, we applied a MinMax scaling that changes the values range to $[-1, 1]$.

We trained multiple machine learning models on the collected fingerprinting dataset and evaluated their performance using a variety of metrics. To improve model performance and evaluation, we introduced additional input features and a new evaluation metric. Specifically, we added the \textit{mean}, \textit{standard deviation (std)}, and \textit{available anchors count} as new features to the input data. Model performance was assessed both with and without these additional features to measure their impact. In addition to conventional classification or regression metrics \textit{MAE} and \textit{MSE}, we introduced a new metric \textbf{Mean Distance Error} \ref{eq:mde} to quantify localization accuracy. This metric calculates the two-dimensional Euclidean distance between the predicted and ground truth coordinates, providing a direct measure of spatial error in meters. The machine learning models evaluated include \textit{Support Vector Machine (SVM)}, \textit{Random Forest}, \textit{Decision Tree}, \textit{Multilayer Perceptron (MLP)}, \textit{Gradient Boosting}, and \textit{Gaussian Process}. These models were selected to represent a diverse set of machine learning paradigms—including kernel-based, ensemble-based, tree-based, neural network-based, and probabilistic approaches—to comprehensively evaluate localization performance under different assumptions and model complexities. The results of these models are presented in the following section.

% \subsubsection{Navigation}
% asd
% Average time for navigation
% Number of trials for testing the navigation algorithm.
% Showing the navigation path and Cost map

\subsection{Vision-based Fallen People Detection}
The discussion that follows centers around the Fallen People dataset, including the cleaning process for the YOLO (You Only Look Once) dataset, and the training of multiple models with various parameters. Key aspects include the computational run times, performance metrics, and constraints related to edge device capabilities.

\subsubsection{Dataset}
For our experiments, we utilized the E-FPDS (Fall Person Detection System) dataset~\citep{maldonado2019fallen}, a publicly available collection specifically designed for fall detection research. Unlike many existing datasets, which often involve simulated environments and limited variations, E-FPDS offers a more realistic and diverse set of images. The dataset contains 6,982 manually labeled RGB images, including 5,023 instances of falls and 2,275 of non-fall activities (e.g., walking, sitting, or lying on a sofa). Each image may include multiple individuals with varying heights (1.2–1.8 meters) and clothing, captured under different lighting conditions and environments. 

All images were acquired using a single fixed camera mounted on a robot at a height of 76 cm above the floor, covering eight distinct environments with variability in perspective, illumination, shadows, and reflections. Each image is accompanied by a .txt annotation file containing bounding box coordinates and a class label indicating fall (1) or non-fall (-1). It is important to note that some images contain no people, while others include both fallen and upright individuals. To provide clarity regarding these cases, we included the table \ref{tab:fallen_count} that summarizes the number of people present in each category.

\begin{table}[t]
\centering
\caption{Distribution of classes across train, validation, and test sets}
\small
\begin{tabular}{lccc}
\toprule
\textbf{Class} & \textbf{Train} & \textbf{Validation} & \textbf{Test} \\
\midrule
Fallen & 3,798 & 761 & 327 \\
Not Fallen & 837 & 410 & 571 \\
No Person & 104 & 0 & 14 \\
Both & 69 & 4 & 61 \\
\midrule
\textbf{Total People} & 4,872 & 1,221 & 1,179 \\
\midrule
\textbf{Total Images} & 4,808 & 1,175 & 973 \\
\bottomrule
\end{tabular}
\label{tab:fallen_count}
\end{table}

To ensure robust model evaluation, we adopted the same data splitting strategy as proposed by the original dataset authors. However, we developed a data cleaning pipeline to restructure the dataset to comply with the YOLO file format. This process included resizing selected images and updating their corresponding bounding box annotations to ensure uniform image dimensions across the dataset. Finally, we converted all bounding box coordinates from absolute pixel values to the normalized format required by YOLO, where positions and dimensions are represented as relative values with respect to image width and height.

\subsubsection{Vision-based Fallen People Detection}

We developed and evaluated two experimental configurations for our fallen-person detection system. In the first configuration, we fine-tuned YOLO models to directly detect and classify fallen individuals without incorporating any auxiliary components. In the second configuration, we implemented a multi-stage pipeline in which a pre-trained YOLO model was employed for initial object detection, followed by the training of a dedicated classifier using the YOLO output as input. This latter approach yielded improved detection performance, as it leveraged a set of heuristic-derived features specifically engineered to enhance the discriminative capacity of the classifier.

YOLO is a state-of-the-art, real-time object detection system that identifies and classifies multiple objects in an image with high speed and accuracy. Unlike traditional approaches that process images in multiple stages, YOLO performs detection in a single forward pass through a convolutional neural network, making it highly efficient for applications such as video surveillance, autonomous driving, and robotics. The algorithm divides the input image into a grid and predicts bounding boxes and class probabilities for each region, facilitating rapid and precise object detection. Over time, YOLO has undergone several iterations, each improving detection accuracy, speed, and adaptability.

\begin{itemize}
    \item \textbf{End-to-End YOLO}: After cleaning the dataset and preparing it for training, we developed a complete pipeline to train and evaluate multiple YOLO networks. Several hyperparameter configurations were tested, and the optimal set was determined empirically through experimentation. We trained the most recent versions of the YOLO architecture—including YOLOv10, YOLOv11, and YOLOv12—resulting in a total of 16 distinct models. The trained models were subsequently transferred to the edge device and converted into TensorRT format to enable efficient real-time inference.

    In this study, the YOLO models were trained to perform object detection with the specific goal of identifying both fallen and upright individuals. Object detection is a core computer vision task in which the model identifies objects within an image and localizes them by drawing bounding boxes.
    
    To evaluate the system’s efficiency and performance, we measured the execution time of various stages, including training, model conversion, model loading, and average inference time, as well as the performance metrics like MAP50, precision, and recall. A comprehensive analysis of these timings, along with the models’ detection performance metrics, is presented in the results section.

    \item \textbf{Multi Stage Detection}: We have used the same data pipeline for this experiment and used the pretrained YOLO models for object detection. We feed all images to the YOLO models and process the outputs for the next model training step. We developed a function to extract and generate meaningful features from the raw bounding boxes out of the YOLO models that can help the detection model for a better understanding of the scene.

    The \textit{extract\_features} function generates a fixed-length feature vector from YOLO-based object detection outputs to support downstream classification. It first filters detected objects to include only relevant classes (e.g., person, chair, bed, couch) and computes the count of each class, total relevant object count, and average width and height of relevant bounding boxes. These features capture scene composition and object scale information.

    For detected persons, the function extracts statistical features based on the vertical center \textit{y\_center} and aspect ratio of bounding boxes, such as mean, standard deviation, max, and min of \textit{y\_center}, and average width-to-height ratio. Additionally, it calculates the minimum Euclidean distance between each person and nearby supportive objects (e.g., bed or chair) to estimate contextual proximity. The resulting vector encodes both scene-level and person-centric spatial features suitable for identifying abnormal postures or fall events.

    Similarly, we used the same YOLO model versions to reach the optimal performance. Also, we trained different classifiers on the generated features for better comparison and performance, including \textit{Random Forest}, \textit{Logistic Regression}, \textit{Gradient Boosting}, and \textit{SVM}. A comprehensive analysis of these timings, along with the models’ detection performance metrics, is presented in the results section.

\end{itemize}

\subsection{End-to-End Evaluation}
In the final evaluation, we prototyped the entire end-to-end functionality of our system by testing it using an experiment performed within our laboratory. The demonstration aimed at ensuring the practical deployability of the intended system. Following Section~\ref{sec:method_overview}, in the online detection phase, the framework continuously acquired IMU sensor readings and applied a trained classifier for identifying the fall event in real-time. Once the trained classifier identifies the fall, the wearable sensor initiates the localization procedure. After the fall location is estimated, a request for robot navigation is sent from the edge device to the robot.

The robot computed a route of travel to the estimated location and, at the same time, started streaming the camera feed to the edge device using ROS 2 messages. We created a specific ROS 2 node on the edge side that subscribes to the camera topic. The node used the learned YOLO model to detect fallen people with two frames per second, as the robot is not moving very fast. Each frame was then annotated with a bounding box around the persons detected, and the annotated images were published by the same node for subsequent processing or visualization.

\begin{figure}
    \centering
  \subfloat[\small{Start}\label{fig:rviz_fallen_1}]{\includegraphics[width=0.95\linewidth]{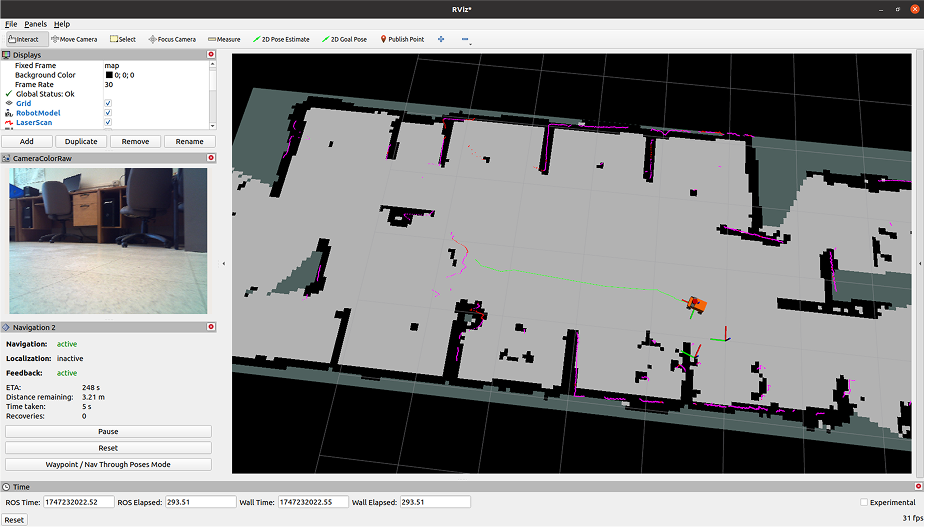}} \\
  \subfloat[\small{Middle}\label{fig:rviz_fallen_2}]{\includegraphics[width=0.95\linewidth]{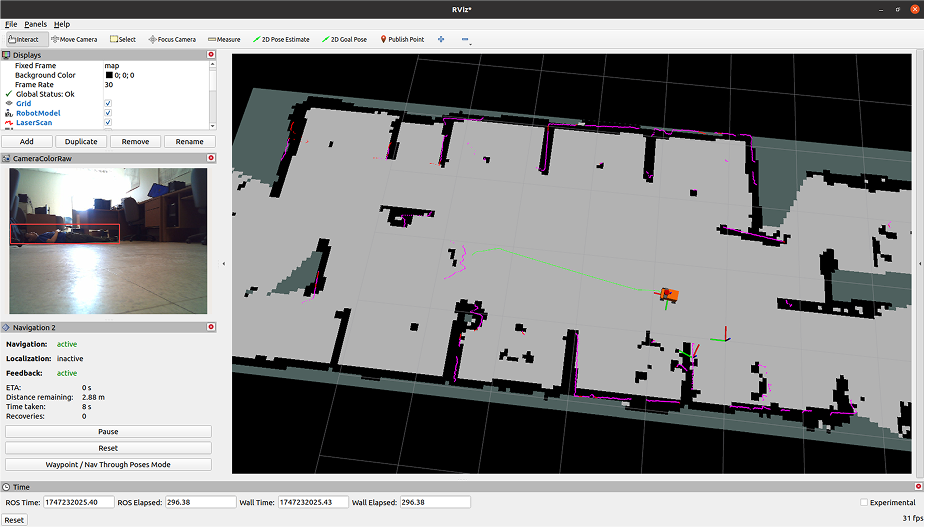}} \\
  \subfloat[\small{End}\label{fig:rviz_fallen_3}]{\includegraphics[width=0.95\linewidth]{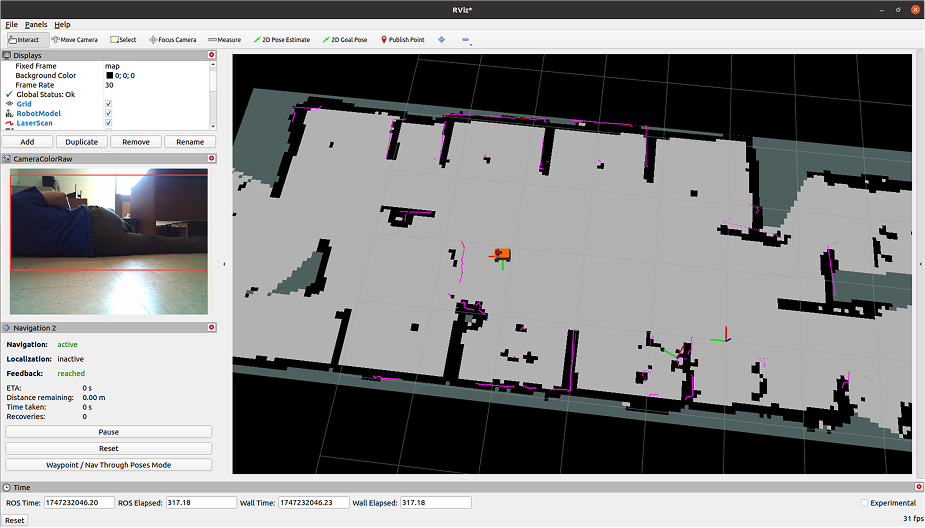}} 
  \centering
     \caption{Three figures from the navigation process from the origin to the location of the fall event}
\label{fig:rviz_fallen}
\end{figure}

\begin{figure}
    \centering
  \subfloat[\small{Start}\label{fig:rviz_not_fallen_1}]{\includegraphics[width=0.95\linewidth]{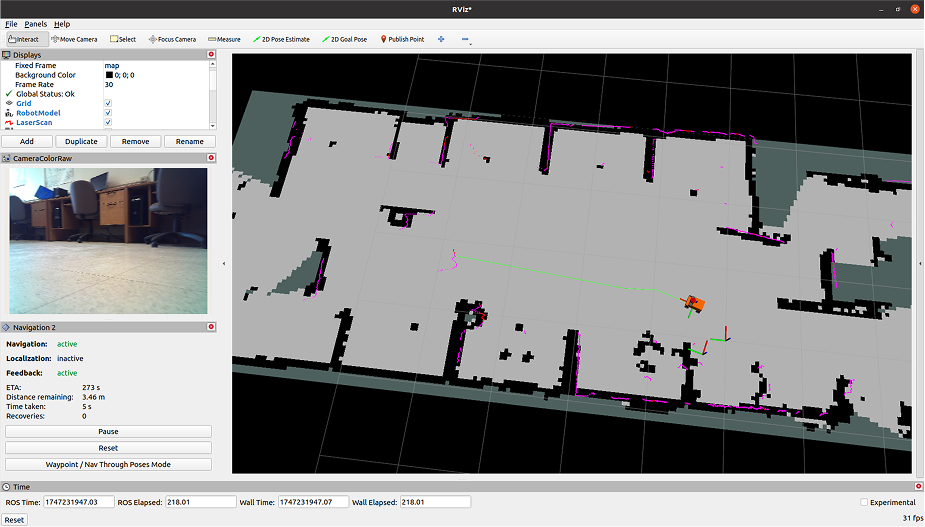}} \\
  \subfloat[\small{Middle}\label{fig:rviz_not_fallen_2}]{\includegraphics[width=0.95\linewidth]{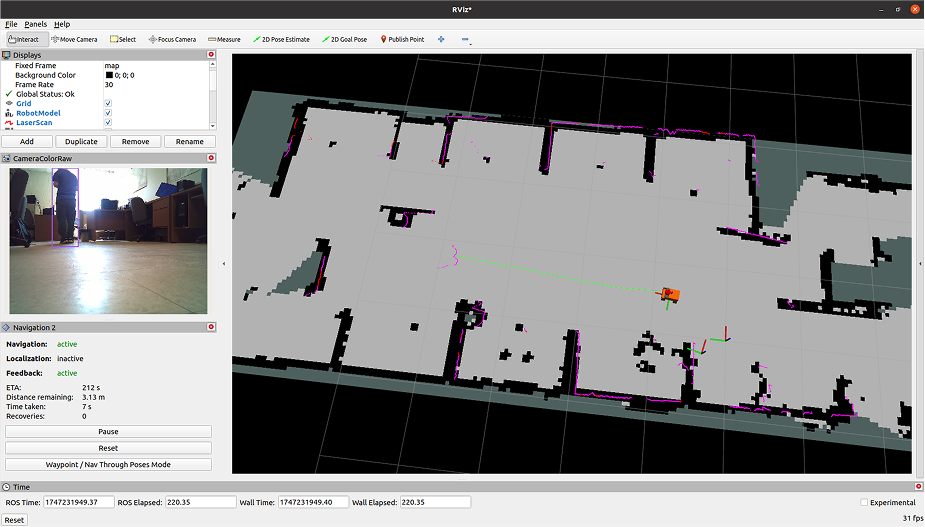}} \\
  \subfloat[\small{End}\label{fig:rviz_not_fallen_3}]{\includegraphics[width=0.95\linewidth]{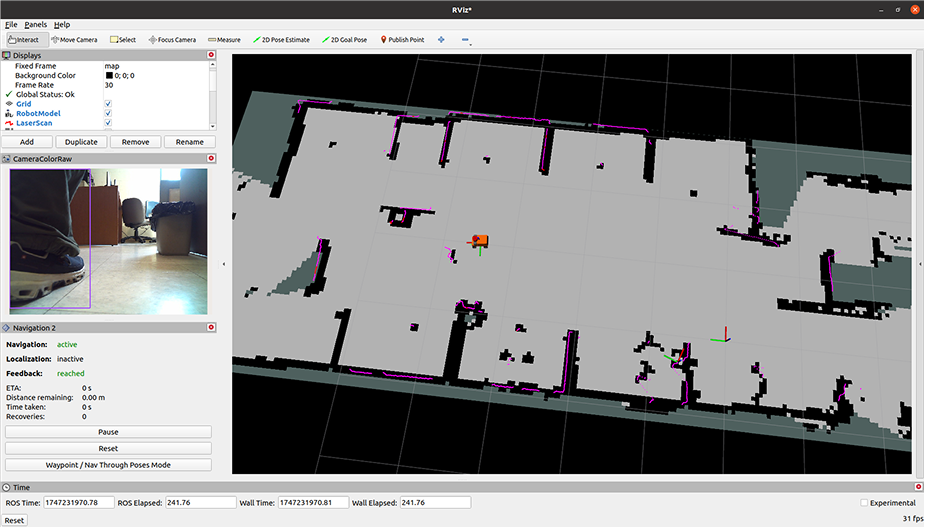}}        
  \caption{Three figures from the navigation process from the origin to the location of the fall event (false alarm).}
        \label{fig:rviz_not_fallen}
\end{figure}

\begin{figure}
    \centering
  \subfloat[\small{Navigation Costmap}\label{fig:rviz_extra_1}]{\includegraphics[width=0.95\linewidth]{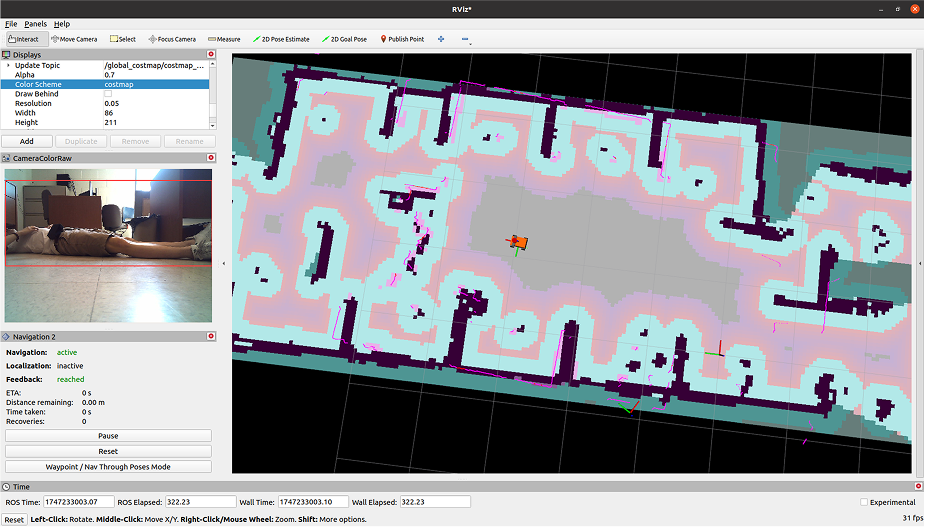}} \\
  \subfloat[\small{Detecting Sitted People As Not Fallen}\label{fig:rviz_extra_2}]{\includegraphics[width=0.95\linewidth]{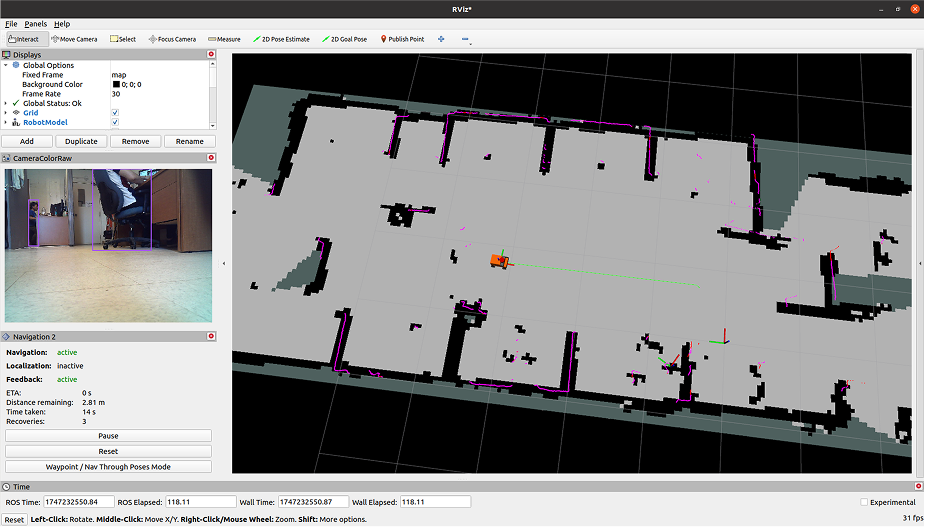}}        
  \caption{Two extra figures for providing more information}
        \label{fig:rviz_extra}
\end{figure}

Last of all, the whole navigation process, camera feed, and the resultant detections could all be watched in real-time using RViz2 as a complete visual interface for system testing. We provided some figures from these experiments that include multiple scenarios. First, in Figure \ref{fig:rviz_fallen}, we provided samples of fallen people where both systems can accurately detect the fall. Second, in the Figure \ref{fig:rviz_not_fallen} we demonstrated a scenario when the semi-supervised federated fall detection raises a false alarm and the robot navigates to the location, but the person has not fallen. Finally, we provided extra images to explain other aspects and possibilities of the framework in Figure \ref{fig:rviz_extra}. 

In Figure \ref{fig:rviz_fallen}, we plotted three screenshots from RViz that were captured from a single scenario. In each image, there was a 2D map on the right side that showed the experimental environment, the robot's location, the navigation planned path, and live lidar scans from the robot's sensor. On the left side, the upper window manages the ROS2 topics that we could see, and the middle window is the live output of the mentioned ROS2 node on the edge device that provided the annotated images with the Yolo model. In Figure \ref{fig:rviz_fallen_2}, the robot is able to detect the fallen people, and there is a red bounding box around the person. Finally, the bottom left window is a navigation plugin that shows the status of the current navigation task. As you can see, the plugin can provide useful information from the nav2 engine, including the passed time, remaining distance, and feedback that shows the task status. In Figure \ref{fig:rviz_fallen_3}, the status is changed to \textit{reached}, which means the robot reached the given destination point.

In Figure \ref{fig:rviz_not_fallen}, we provided three continuous screenshots from RViz that were captured from another scenario. We provided fake information to the semi-supervised federated fall detection model to trigger the fall event procedure, but the person who is wearing the wearable device is standing. The robot navigated to the wearable device's estimated location and found a person who was standing. As you can see in the Figure \ref{fig:rviz_not_fallen_2} and Figure \ref{fig:rviz_not_fallen_3}, the bounding box color is not red, which means the fallen people detection model identifies the person as not fallen.

Figure \ref{fig:rviz_extra_1} shows the live navigation costmap, which was used by nav2 for path planning and navigation, and it was updated with the live data from the lidar sensor. The live costmap could prevent the robot from hitting obstacles or the fallen person. Also, Figure \ref{fig:rviz_extra_2} showed that the model identified two people who are sitting as not fallen, which is a common false alarm from the signal-based fall detection systems. 

\section{Results}
\label{sec:results}
We evaluated our multi-stage fall detection framework through a series of real-world experiments, assessing each component—fall detection, localization and navigation, and vision-based confirmation. Both quantitative metrics and qualitative scenarios are presented to demonstrate the system’s overall accuracy, robustness, and readiness for real-time deployment in elder care settings.

\subsection{Semi-Supervised Federated Fall Detection}
In this section, we present and analyze the results of our proposed Semi-Supervised Federated Fall Detection framework. We compare our methods against a recent state-of-the-art supervised approach on the same dataset. Table \ref{tab:sf2d_results} summarizes the quantitative evaluation using standard classification metrics: Accuracy (ACC), Precision (PR), Recall (RE), and F1 Score (F1). These metrics are formally defined below, followed by an in-depth discussion of the results.

\begin{table}[ht]
    \centering
    \caption{Comparison of different methods (ACC = Accuracy, PR = Precision, RE = Recall, and F1 = F1 score).}
    \captionsetup{width=\textwidth}
    \footnotesize
    \begin{tabular}{cccccc}
        \toprule
        \textbf{Method} & \textbf{ACC}$\uparrow$ & \textbf{PR}$\uparrow$ & \textbf{RE}$\uparrow$ &  \textbf{F1}$\uparrow$ \\
        \midrule
        Fed-ELM$^\dagger$ \citep{yu_elderly_2022} & 98.01 & - & 95.25 & - \\
        \midrule
        CL (Ours)$^\ddagger$ & \textbf{99.33} & 99.67 & \textbf{99.63} & \textbf{99.65} \\
        \textbf{FL (Ours)$^\ddagger$} & 99.19 & \textbf{99.69} & 99.47 & 99.58 \\
        \bottomrule
    \end{tabular}\\
    \footnotesize{$^\dagger$ Supervised $^\ddagger$ Semi-supervised}
    \label{tab:sf2d_results}
\end{table}

The evaluation metrics are defined as follows. Let $TP$, $TN$, $FP$, and $FN$ denote true positives, true negatives, false positives, and false negatives, respectively.

\begin{itemize}
    \item \textbf{Accuracy (ACC)} measures the overall correctness of the model:
    \begin{equation}
    \text{ACC} = \frac{TP + TN}{TP + TN + FP + FN}
    \label{eq:accuracy}
    \end{equation}

    \item \textbf{Precision (PR)} indicates how many of the predicted positive instances are actually positive:
    \begin{equation}
    \text{PR} = \frac{TP}{TP + FP}
    \label{eq:precision}
    \end{equation}

    \item \textbf{Recall (RE)} measures how many of the actual positive instances were correctly identified:
    \begin{equation}
    \text{RE} = \frac{TP}{TP + FN}
    \label{eq:recall}
    \end{equation}

    \item \textbf{F1 Score (F1)} is the harmonic mean of precision and recall:
    \begin{equation}
    \text{F1} = 2 \times \frac{\text{PR} \times \text{RE}}{\text{PR} + \text{RE}}
    \label{eq:f1}
    \end{equation}
\end{itemize}

These metrics offer a comprehensive understanding of the model’s classification performance, especially in scenarios like fall detection, where both false positives and false negatives carry significant consequences.

The baseline method, Fed-ELM \citep{yu_elderly_2022}, achieved an accuracy of 98.01\% and a recall of 95.25\%, but did not report precision or F1 score. In contrast, our proposed semi-supervised approaches significantly outperformed this baseline across all reported metrics.

The \textbf{Centralized Learning (CL)} variant of our method yielded the highest overall accuracy (99.33\%), along with high precision (99.67\%), recall (99.63\%), and F1 score (99.65\%). These results indicate that access to the complete dataset during training, even in a semi-supervised setting, allows the model to learn robust representations and accurately detect fall events with minimal false positives or missed detections.

The \textbf{Federated Learning (FL)} variant, while operating under data-sharing constraints and limited label availability, achieved comparable performance: 99.19\% accuracy, the highest precision (99.69\%), 99.47\% recall, and a strong F1 score of 99.58. These results demonstrate the viability of semi-supervised federated learning for real-world deployments, offering a balance between privacy preservation and high detection performance.

Interestingly, the FL approach outperformed CL in precision, suggesting that the federated training process may promote generalization and reduce overfitting to specific data distributions. This characteristic is particularly important in decentralized and heterogeneous environments such as smart homes or elder care facilities.

Our findings confirm that semi-supervised learning, when combined with either centralized or federated paradigms, can exceed the performance of fully supervised models in fall detection. The federated setup, in particular, presents a compelling solution for privacy-aware applications, achieving nearly the same effectiveness as its centralized counterpart without requiring raw data exchange. These results highlight the practical potential of our approach in enabling accurate, privacy-preserving fall detection across distributed edge devices.

\subsection{Fingerprinting Dataset Creation}

In this section, we evaluate the effectiveness of our proposed automated fingerprinting dataset creation approach in comparison with several existing techniques. The comparison considers three essential dimensions for practical deployment: time efficiency, surveying platform, and odometry method. The summarized results are presented in Table \ref{tab:mapping_results}, highlighting the scalability and performance of each approach.

\begin{table*}[t!]
\centering
\caption{Effectiveness evaluation of the proposed approach in comparison with other approaches}
\small
\begin{tabular}{lccc}
\toprule
\textbf{Method} & \begin{tabular}{@{}c@{}} \textbf{Time Efficiency}$\uparrow$ \\ (\textit{Reference Points per Second}) \end{tabular} & \textbf{Surveying Platform} & \textbf{Odometry Method} \\
\midrule
Ours & \textbf{1.00}  & Robot  & LiDAR SLAM  \\
\citet{abu_kharmeh_indoor_2023} & 0.14 & Robot & Black Tape \\
\citet{rizk_laser_2023} & n/a & Crowdsourcing (individuals) & Laser-Range Scan Tracking\\ 
\citet{silva_industrial_2023} & 0.62 & Manually Pushed Trolley & ArUco Tags with Camera\\ 
\bottomrule
\end{tabular}
\label{tab:mapping_results}
\end{table*}

Our method demonstrates the highest time efficiency, producing \textbf{1 reference point per second}, a substantial improvement over other approaches. Compared to \citet{abu_kharmeh_indoor_2023}, which achieves only 0.14 reference points per second, our system is over \textbf{7 times faster}. Even when compared with more efficient systems like \citet{silva_industrial_2023} (0.62 reference points/sec), our approach offers a significant performance edge, resulting in faster and more scalable data collection.

Unlike manual or crowdsourced platforms, our method uses a fully autonomous mobile robot, enabling consistent and repeatable data collection across different indoor environments. In contrast, \citet{silva_industrial_2023} relied on a manually pushed trolley, and \citet{rizk_laser_2023} used crowdsourcing by individuals, both of which introduce user variability and logistical complexity.

We leverage a LiDAR-based SLAM approach for localization, offering robust and accurate trajectory tracking, even in complex or dynamic environments. In contrast, other methods, such as black-tape following \citet{abu_kharmeh_indoor_2023} or visual tracking with ArUco tags \citep{silva_industrial_2023}, may suffer from visual obstructions or environmental constraints. Our method ensures reliable motion estimation without requiring high-contrast visual markers or specialized surface conditions.

While our method eliminates the need for physical modifications like tape or tags, it does require the placement of BLE anchors to enable signal-based fingerprinting. This setup step is a one-time infrastructure requirement but enables long-term scalability and automation. Unlike visual or tactile preparation methods, BLE anchors are minimally intrusive and easily maintained, making them suitable for semi-permanent deployments.

Our approach offers a high-performance, robot-driven, and infrastructure-light solution for automated fingerprinting. It surpasses existing methods in data collection speed and odometry robustness, while maintaining operational flexibility across indoor spaces. The balance between initial BLE anchor deployment and long-term automation benefits makes it a highly viable solution for real-world fingerprinting at scale.

\subsection{Localization Model}

In this section, we evaluate the performance of several machine learning models for indoor localization using fingerprinting data. We specifically examine the impact of feature engineering on model accuracy by comparing models trained with and without newly derived features. The evaluation is based on three standard regression metrics: Mean Absolute Error (MAE), Mean Squared Error (MSE), and Mean Distance Error (MDE), as presented in Table \ref{tab:localization_result}.

\begin{table}[ht]
\scriptsize  % Reduce font size to help it fit
\centering
\caption{Performance Comparison of Models With and Without Feature Engineering. \scriptsize{The best result is in \textbf{bold}, and the second best result is \underline{underlined}.}}
\renewcommand{\arraystretch}{1.2}
\begin{tabular}{|l@{\hskip 3pt}|c@{\hskip 3pt}c@{\hskip 3pt}c@{\hskip 3pt}|c@{\hskip 3pt}c@{\hskip 3pt}c@{\hskip 3pt}|}
    \hline
    \multirow{2}{*}{\textbf{Model}} & \multicolumn{3}{c|}{\textbf{Without New Features}} & \multicolumn{3}{c|}{\textbf{With New Features}} \\
    \cline{2-7}
     & \textbf{MAE}$\downarrow$ & \textbf{MSE}$\downarrow$ & \textbf{MDE}$\downarrow$ & \textbf{MAE}$\downarrow$ & \textbf{MSE}$\downarrow$ & \textbf{MDE}$\downarrow$ \\
    \hline
    SVM              & 0.6938 & 0.9656  & 1.1134 & 0.7022 & 0.9821  & 1.1231 \\
    Random Forest    & \textbf{0.6757} & \textbf{0.9022}  & \textbf{1.0700} & \textbf{0.6240} & \textbf{0.7843}  & \textbf{0.9873} \\
    Decision Tree    & 0.8863 & 1.7553  & 1.4075 & 0.8503 & 1.7640  & 1.3962 \\
    MLP              & 0.7252 & 0.9621  & 1.1552 & 0.6935 & 0.9603  & 1.1209 \\
    Gradient Boosting& \underline{0.6867} & \underline{0.8888}  & \underline{1.1119} & \underline{0.6772} & \underline{0.8464}  & \underline{1.0798} \\
    Gaussian Process & 3.9781 & 21.9609 & 6.2679 & 3.2834 & 19.7971 & 5.8471 \\
    \hline
\end{tabular}
\label{tab:localization_result}
\end{table}

The three metrics used for model evaluation are defined as follows:
\begin{itemize}
    \item \textbf{Mean Absolute Error (MAE)}: Measures the average magnitude of errors between predicted and actual locations, without considering direction.
    \begin{equation}
    \text{MAE} = \frac{1}{n} \sum_{i=1}^{n} |y_i - \hat{y}_i|
    \label{eq:mae}
    \end{equation}
    
    \item \textbf{Mean Squared Error (MSE)}: Penalizes larger errors more than MAE by squaring the differences.
    \begin{equation}
    \text{MSE} = \frac{1}{n} \sum_{i=1}^{n} (y_i - \hat{y}_i)^2
    \label{eq:mse}
    \end{equation}
    
    \item \textbf{Mean Distance Error (MDE)}: Represents the average Euclidean distance between predicted and actual spatial coordinates.
    \begin{equation}
    \text{MDE} = \frac{1}{n} \sum_{i=1}^{n} \sqrt{(y_{i1} - \hat{y}_{i1})^2 + (y_{i2} - \hat{y}_{i2})^2}
    \label{eq:mde}
    \end{equation}
\end{itemize}
Feature engineering demonstrated a generally positive impact across most models. Notably, the \textbf{Random Forest} model exhibited the greatest improvement with the introduction of new features, achieving the lowest error values across all metrics: MAE (0.6240), MSE (0.7843), and MDE (0.9873). These results indicate improved model generalization and finer granularity in location predictions.

\textbf{Gradient Boosting} and \textbf{MLP} also benefited from feature engineering, showing modest yet consistent reductions in all error metrics. For example, Gradient Boosting improved from an MSE of 0.8888 to 0.8464 and MDE from 1.1119 to 1.0798, indicating enhanced stability in high-density feature spaces.

Interestingly, \textbf{SVM} showed a slight degradation in performance after feature engineering, which may reflect overfitting or sensitivity to irrelevant features. Similarly, \textbf{Decision Tree} performance improved only marginally, suggesting limited capacity to benefit from added feature complexity.

The \textbf{Gaussian Process} model yielded the poorest results, with high error rates in both configurations. While its performance improved after feature engineering (e.g., MDE reduced from 6.2679 to 5.8471), it remained unsuitable for this task due to scalability issues and poor generalization in high-dimensional input spaces.

The results validate the effectiveness of feature engineering for enhancing indoor localization accuracy, particularly for ensemble-based models like Random Forest and Gradient Boosting. The Random Forest model, when combined with well-designed features, offers a highly accurate and robust solution for fingerprint-based localization. This highlights the importance of tailored feature construction in location-aware systems and supports the deployment of machine learning models in real-world indoor positioning applications.

\subsection{Vision-based Fallen People Detection}
In the following section, we present a comprehensive evaluation of the Vision-based Fallen People Detection. First, we assess the performance of various YOLO models through a comparative analysis. Second, we report the results of the two-stage classification process, employing multiple classifiers to enhance detection accuracy.

\subsubsection{End-to-End YOLO}
We evaluate the system on various performance metrics, including execution time and detection accuracy. Additionally, we analyze the trade-offs between model performance and efficiency, particularly focusing on inference speed and detection precision.
\begin{itemize}
    \item \textbf{Metrics}: 
    For evaluating the performance of the trained YOLO models, we utilized several metrics: inference time, mean Average Precision at 50\% Intersection over Union (mAP50), precision, and recall. To obtain more reliable results, we performed inference on multiple samples and averaged the times for each stage of the process. 

    The \textit{mean Average Precision at 50\% Intersection over Union} (mAP50) is computed as the average of the precision at each recall level, where the Intersection over Union (IoU) threshold is set to 50\%. This metric provides a measure of the model’s ability to detect objects accurately and is given by:

    \begin{equation}
    \text{mAP50} = \frac{1}{N} \sum_{i=1}^{N} \text{AP}_{i}(0.5)
    \label{eq:map}
    \end{equation}

    where \( \text{AP}_i(0.5) \) denotes the Average Precision at IoU threshold of 50\% for the \(i\)-th class, and \(N\) is the total number of classes.

    Precision and recall are key metrics for evaluating the quality of the detections. \textbf{Precision} measures the proportion of true positive detections among all positive predictions, and is given by Equation \ref{eq:precision}, where \( TP \) is the number of true positives and \( FP \) is the number of false positives.
    
    \textbf{Recall}, on the other hand, measures the proportion of true positive detections among all ground truth instances, and is given by Equation \ref{eq:recall}, where \( FN \) is the number of false negatives. These metrics help assess the balance between detecting all objects (recall) and ensuring that the detected objects are correct (precision).

    For the inference time, we averaged the time taken for each inference across multiple samples to account for variability. This average inference time represents the model's efficiency in processing inputs in a real-world deployment setting. The total inference time for each model is computed as:
    
    \begin{equation}
    \text{Average Inference Time} = \frac{1}{M} \sum_{j=1}^{M} T_j
    \label{eq:ait}
    \end{equation}
    
    where \( M \) is the total number of test samples, and \( T_j \) is the time taken to process the \(j\)-th sample. These metrics were used to provide a comprehensive evaluation of the YOLO models, considering both their accuracy and efficiency in detecting fallen and upright individuals.
    
    \item \textbf{Training Time vs. mAP50}:
    Figure~\ref{fig:training_time_vs_map_bubble} illustrates the relationship between the training time and mAP50 for each model. The x-axis represents the training time in seconds, while the y-axis shows the mAP50 score, which is a measure of the model's accuracy. Each model is represented by a circle, where the size of the circle corresponds to the model's size in megabytes. Larger circles indicate models with more parameters and a higher memory footprint. From the plot, we can observe that there is no clear linear relationship between training time and mAP50. While some models with longer training times, such as \texttt{YOLOv10b}, yield higher mAP50 scores, other models, like \texttt{YOLOv10n}, have relatively shorter training times with higher mAP50 values, indicating more efficient training.

    \begin{figure}[htb] 
        \centering
        \includegraphics[width=0.95\linewidth]{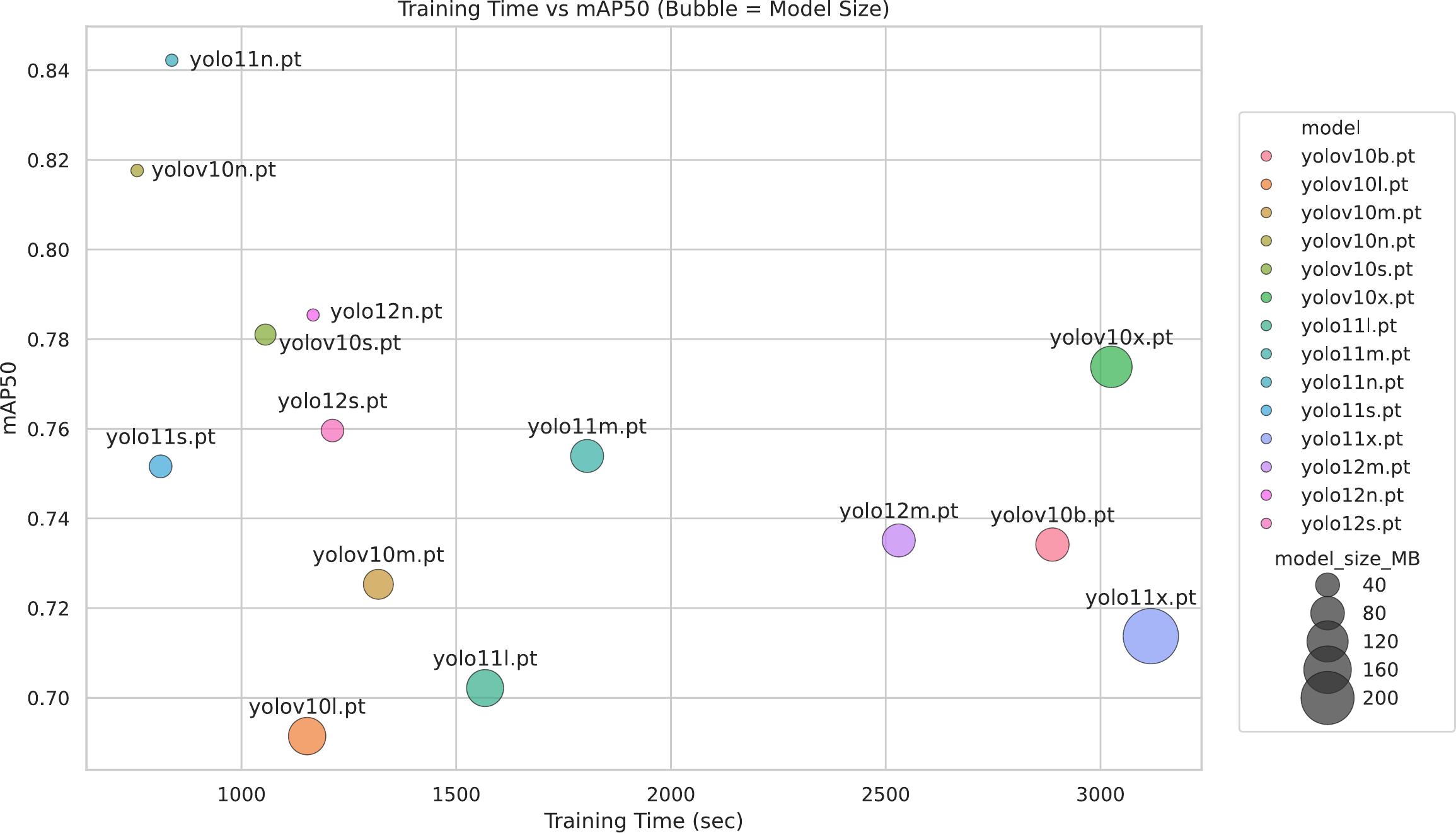}
        \caption{mAP50 vs Training time vs Size}
        \label{fig:training_time_vs_map_bubble}
    \end{figure}

    \item \textbf{mAP50 vs. Inference Time}:
    Figure~\ref{fig:map_vs_inference_time} shows the trade-off between mAP50 and inference time for each model. The x-axis represents the average inference time in seconds, while the y-axis shows the mAP50 score. This plot highlights the balance between model accuracy and inference speed. Models with faster inference times, such as \texttt{YOLOv10n}, tend to have higher mAP50 values, while models with slower inference times, like \texttt{YOLOv10x}, have lower accuracy scores. This suggests that there is a direct impact of inference time on detection performance.

    \begin{figure}[htb] 
        \centering
        \includegraphics[width=0.95\linewidth]{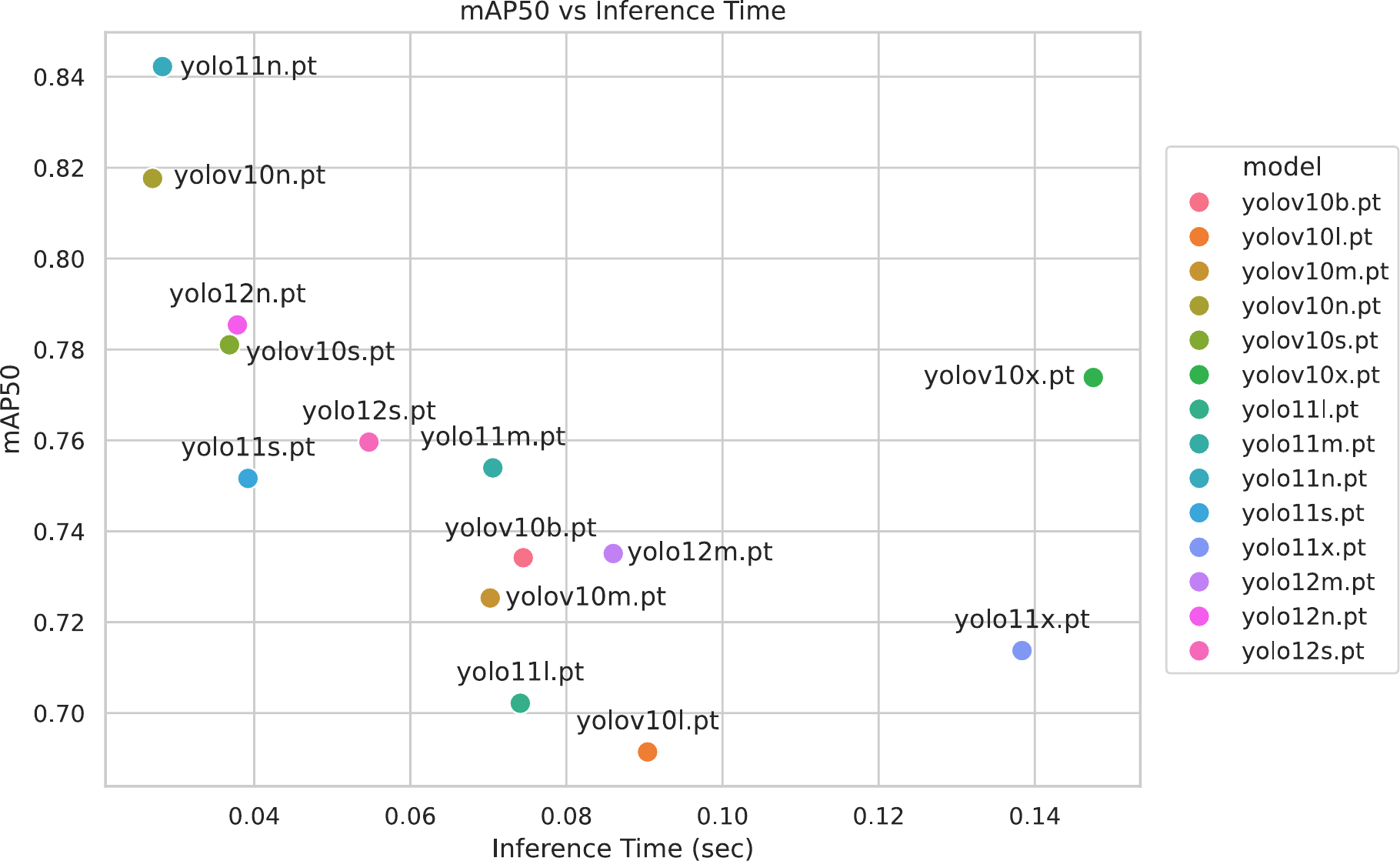}
        \caption{mAP50 vs Inference time}
        \label{fig:map_vs_inference_time}
    \end{figure}

    \item \textbf{mAP50 vs. Model Size}:
    Figure~\ref{fig:map_vs_model_size} explores the relationship between model size and mAP50. The x-axis indicates the model size in megabytes, and the y-axis shows the mAP50 score. This figure demonstrates that, in general, larger models tend to have better detection performance (higher mAP50), as seen with models such as \texttt{YOLOv10b} and \texttt{YOLOv10x}. However, smaller models, like \texttt{YOLOv10n}, also achieve high mAP50 values despite their smaller size, highlighting that model size is not the only determinant of accuracy.

    \begin{figure}[htb] 
        \centering
        \includegraphics[width=0.95\linewidth]{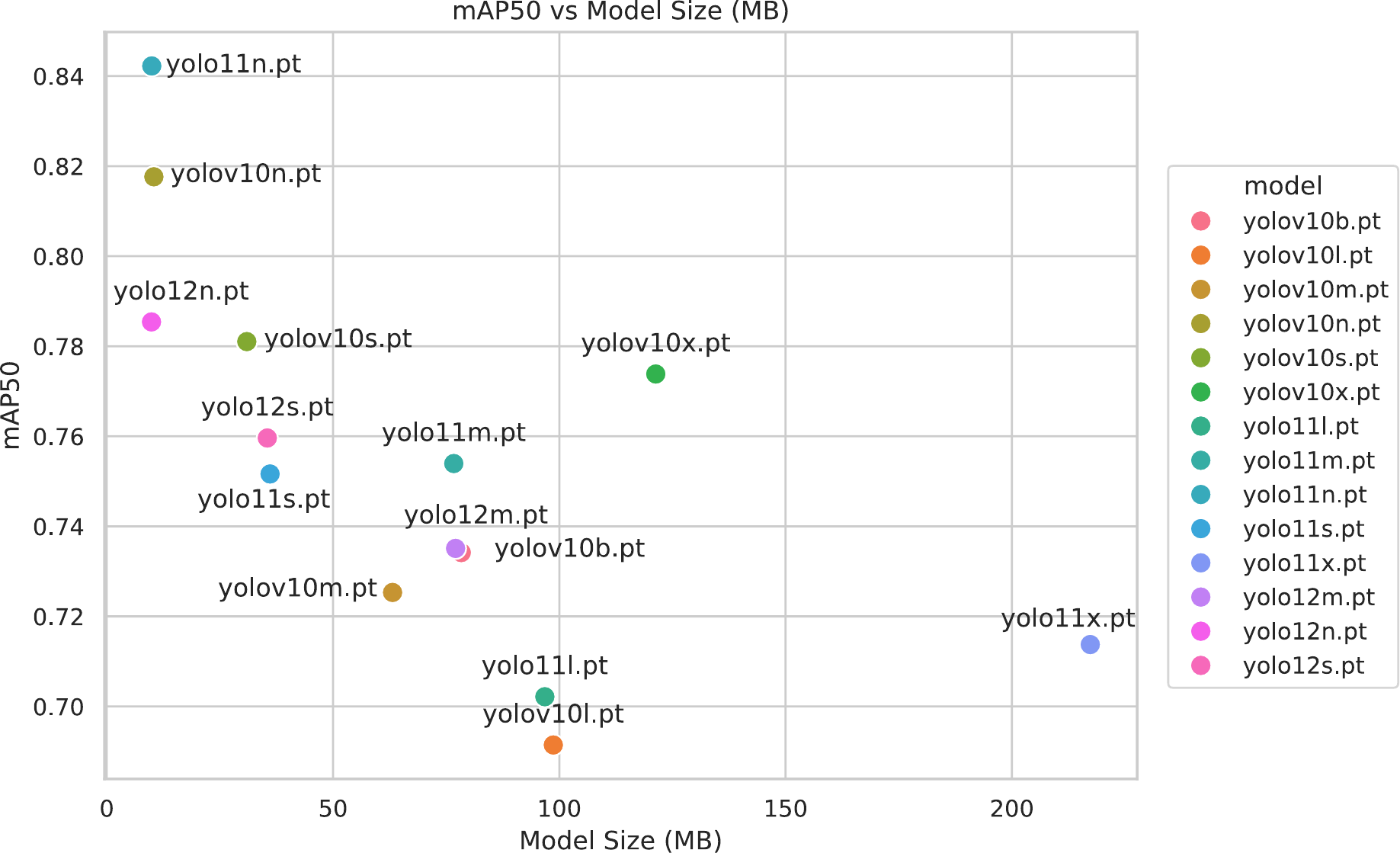}
        \caption{mAP50 vs Model Size}
        \label{fig:map_vs_model_size}
    \end{figure}

    \item \textbf{Comparison of mAP50 for All Models}:
    Figure~\ref{fig:map50_rank} presents a bar chart comparing the mAP50 scores of all models. The bars are sorted in descending order, allowing for an easy comparison of model performance. From the chart, it is clear that the models with higher mAP50 scores, such as \texttt{YOLOv10n} and \texttt{YOLO11n}, deliver better detection accuracy compared to others. On the other hand, models like \texttt{YOLOv10l} and \texttt{YOLO11x} exhibit lower mAP50 values, suggesting a trade-off between accuracy and other factors such as model size and inference time.

    \begin{figure}[htb] 
        \centering
        \includegraphics[width=0.95\linewidth]{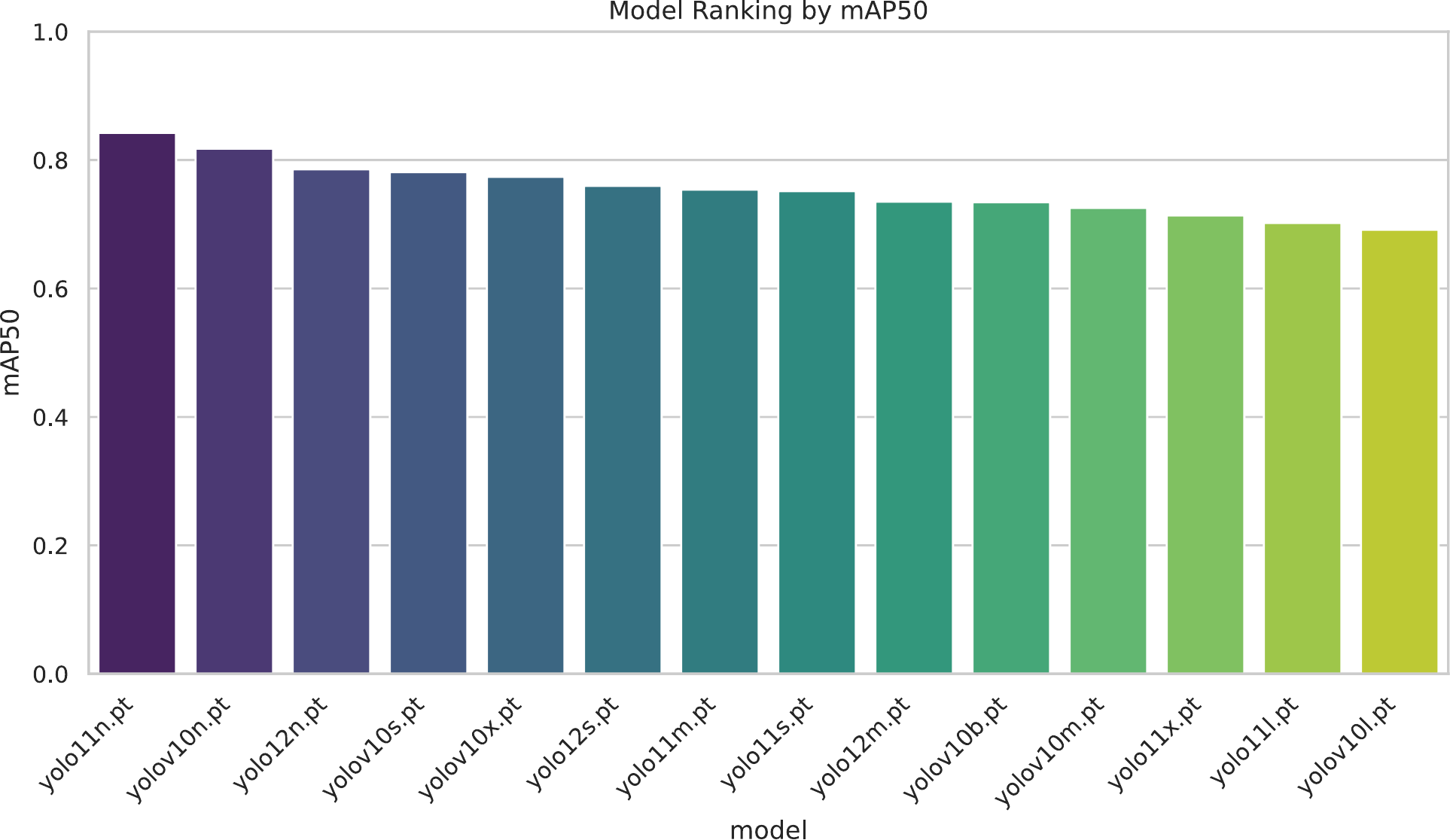}
        \caption{Trained Yolo models sorted based on MAP50 metric}
        \label{fig:map50_rank}
    \end{figure}

    \item \textbf{Model Comparison Table}:
    Table~\ref{tab:yolo_performance} summarizes the key characteristics of each model, including training time, model size, mAP50, inference time, and other relevant times. This table allows for a detailed comparison of the models based on various performance metrics.

    \begin{table*}[ht]
    \small
    \centering
    \caption{Summary of model characteristics including training time, model size, mAP50, and inference times. \scriptsize{The best result is in \textbf{bold}, and the second best result is \underline{underlined}.}}
    \begin{tabular}{lrrrrrr}
    \toprule
    \textbf{Model} & 
    \shortstack{\textbf{Training Time}\\(sec)$\downarrow$} & 
    \shortstack{\textbf{Model Size}\\(MB)$\downarrow$} & 
    \shortstack{\textbf{mAP50}$\uparrow$} & 
    \shortstack{\textbf{Inference Time}\\(sec)$\downarrow$} & 
    \shortstack{\textbf{Model Loading}\\Time(sec)$\downarrow$} & 
    \shortstack{\textbf{Export}\\\textbf{Time}(sec)$\downarrow$} \\
    \midrule
    yolov10b & 2888.088 & 78.331 & 0.734 & 0.074 & 0.001 & 898.422 \\
    yolov10l & 1152.831 & 98.674 & 0.691 & 0.090 & 0.001 & 1027.874 \\
    yolov10m & 1318.723 & 63.149 & 0.725 & 0.070 & 0.001 & 625.399 \\
    yolov10n & \textbf{757.104} & 10.407 & \underline{0.818} & \textbf{0.027} & 0.001 & 237.979 \\
    yolov10s & 1055.842 & 30.939 & 0.781 & 0.037 & 0.001 & 410.928 \\
    yolov10x & 3025.381 & 121.310 & 0.774 & 0.147 & 0.001 & 1207.830 \\
    yolo11l & 1567.219 & 96.797 & 0.702 & 0.074 & 0.002 & 840.390 \\
    yolo11m & 1804.687 & 76.678 & 0.754 & 0.071 & 0.002 & 747.258 \\
    yolo11n & 837.733 & \underline{9.940} & \textbf{0.842} & \underline{0.028} & 0.002 & \textbf{203.824} \\
    yolo11s & \underline{811.772} & 36.082 & 0.752 & 0.039 & 0.002 & 385.173 \\
    yolo11x & 3117.215 & 217.323 & 0.714 & 0.138 & 0.002 & 1001.113 \\
    yolo12m & 2530.269 & 77.074 & 0.735 & 0.086 & 0.001 & 676.406 \\
    yolo12n & 1166.732 & \textbf{9.885} & 0.785 & 0.038 & 0.001 & \underline{228.748} \\
    yolo12s & 1211.850 & 35.472 & 0.760 & 0.055 & 0.002 & 357.168 \\
    \bottomrule
    \end{tabular}
    \label{tab:yolo_performance}
    \end{table*}

    In addition to the mAP50 scores, we also evaluate the models based on precision and recall, which are crucial for assessing the models' ability to correctly identify and localize the fallen and upright individuals. Precision represents the proportion of true positive detections among all the predicted positives, while recall measures the proportion of true positive detections among all actual instances of the objects in the ground truth. Table~\ref{tab:precision_recall} presents the precision, recall, and mAP50 scores for all models.

    As seen in the table, the model \texttt{YOLOv10n} exhibits the highest precision (0.815) and recall (0.754), which also results in the highest mAP50 score (0.828), making it the best-performing model in terms of overall detection accuracy. The model \texttt{YOLO11n} also demonstrates a strong performance with precision (0.801) and recall (0.818), and a similarly high mAP50 of 0.842, indicating it is highly capable of detecting both fallen and upright individuals. On the other hand, models like \texttt{YOLOv10x} and \texttt{YOLO12s} show relatively lower precision and recall values, with \texttt{YOLOv10x} achieving the lowest precision score of 0.746, which corresponds to its lower mAP50 of 0.774. Overall, the precision and recall values provide further insights into the models' trade-offs between detecting as many objects as possible (high recall) and ensuring the detections are accurate (high precision).

    \begin{table}[ht]
    \small  % Reduce font size to help it fit
    \centering
    \caption{Precision, recall, and mAP50 scores for all YOLO models (PR = Precision, RE = Recall). \scriptsize{The best result is in \textbf{bold}, and the second best result is \underline{underlined}.}}
    \begin{tabular}{lrrr}
    \toprule
    \textbf{Model} & \textbf{PR}$\uparrow$ & \textbf{RE}$\uparrow$ & \textbf{mAP50}$\uparrow$ \\
    \midrule
    yolov10b & 0.789 & 0.751 & 0.769 \\
    yolov10l & 0.797 & 0.650 & 0.707 \\
    yolov10m & 0.784 & 0.659 & 0.733 \\
    yolov10n & \textbf{0.815} & 0.754 & \underline{0.828} \\
    yolov10s & 0.801 & 0.738 & 0.781 \\
    yolov10x & 0.746 & 0.750 & 0.774 \\
    yolo11l & 0.752 & 0.661 & 0.702 \\
    yolo11m & 0.789 & 0.729 & 0.760 \\
    yolo11n & \underline{0.801} & \textbf{0.818} & \textbf{0.842} \\
    yolo11s & 0.776 & \underline{0.794} & 0.795 \\
    yolo11x & 0.776 & 0.674 & 0.714 \\
    yolo12m & 0.758 & 0.709 & 0.734 \\
    yolo12n & 0.769 & 0.785 & 0.790 \\
    yolo12s & 0.710 & 0.728 & 0.761 \\
    \bottomrule
    \end{tabular}
    \label{tab:precision_recall}
    \end{table}
    
    \item \textbf{Discussion}
    The analysis of the figures and tables provides valuable insights into the trade-offs between model performance (mAP50), training time, model size, and inference time. While some smaller models, such as \texttt{YOLOv10n} and \texttt{yolo11n}, exhibit superior mAP50 scores with fast inference times, larger models like \texttt{YOLOv10b} and \texttt{YOLOv10x} tend to achieve lower accuracy but are slower in terms of inference. These results suggest that the choice of model depends on the specific application requirements, where faster inference may be prioritized over higher accuracy, or vice versa. The sorted bar chart of mAP50 in Figure~\ref{fig:map50_rank} highlights the models that provide the best detection accuracy, while the other figures demonstrate the performance trade-offs in real-world scenarios.
    
\end{itemize}

\subsubsection{Multi Stage Detection}

In this experiment, we utilized a data pipeline for object detection, applying pretrained YOLO models and extracting meaningful features from the YOLO outputs to aid downstream classification tasks. The goal of this experiment was to compare the performance of various classifiers trained on features derived from the YOLO model outputs and to determine the best combination of YOLO models and classifiers.

% \textbf{Evaluation Metrics:} The evaluation of classifier performance is based on several key metrics, which are commonly used in classification tasks:

% - \textbf{Precision}: Precision measures the proportion of true positive predictions out of all positive predictions. It is defined as:
%   \[
%   \text{Precision} = \frac{TP}{TP + FP}
%   \]
%   where \( TP \) is the number of true positives and \( FP \) is the number of false positives.

% - \textbf{Recall (Sensitivity)}: Recall measures the proportion of true positives out of all actual positive instances. It is given by:
%   \[
%   \text{Recall} = \frac{TP}{TP + FN}
%   \]
%   where \( FN \) is the number of false negatives.

% - \textbf{F1-Score}: The F1-score is the harmonic mean of precision and recall, providing a balanced measure of a model’s performance, particularly when there is an imbalance between classes. It is calculated as:
%   \[
%   \text{F1-Score} = 2 \cdot \frac{\text{Precision} \cdot \text{Recall}}{\text{Precision} + \text{Recall}}
%   \]
  
% - \textbf{Accuracy}: Accuracy measures the overall proportion of correct predictions. It is calculated as:
%   \[
%   \text{Accuracy} = \frac{TP + TN}{TP + TN + FP + FN}
%   \]
%   where \( TN \) is the number of true negatives.

% Each of these metrics provides different insights into the model's ability to correctly classify instances, and they are especially useful when evaluating models in multi-class or imbalanced classification problems.

\textbf{Evaluation Metrics:} The evaluation of classifier performance is based on several key metrics, which are commonly used in classification tasks:

\begin{itemize}
    \item \textbf{Precision}: Precision measures the proportion of true positive predictions out of all positive predictions. It is given by Equation \ref{eq:precision}, where \( TP \) is the number of true positives and \( FP \) is the number of false positives.
    
    \item \textbf{Recall (Sensitivity)}: Recall measures the proportion of true positives out of all actual positive instances. It is given by Equation \ref{eq:recall}, where \( FN \) is the number of false negatives.
    
    \item \textbf{F1-Score}: The F1-score is the harmonic mean of precision and recall, providing a balanced measure of a model’s performance, particularly when there is an imbalance between classes. It is calculated using the formula in Equation \ref{eq:f1}.
    
    \item \textbf{Accuracy}: Accuracy measures the overall proportion of correct predictions. It is calculated as shown in Equation \ref{eq:accuracy}, where \( TN \) is the number of true negatives.
\end{itemize}

Each of these metrics provides different insights into the model's ability to correctly classify instances, and they are especially useful when evaluating models in multi-class or imbalanced classification problems.

\textbf{Classifier Performance Comparison:} The first set of results focuses on the performance of different classifiers, including \texttt{Random Forest}, \texttt{Logistic Regression}, \texttt{Gradient Boosting}, and \texttt{SVM}, evaluated using various performance metrics such as precision, recall, F1-score, and accuracy. 

\begin{figure*}[t] 
    \centering
  \subfloat[\small{PR and AUC for top 3 Classifiers}\label{fig:vfpd_auc}]{\includegraphics[width=0.95\linewidth]{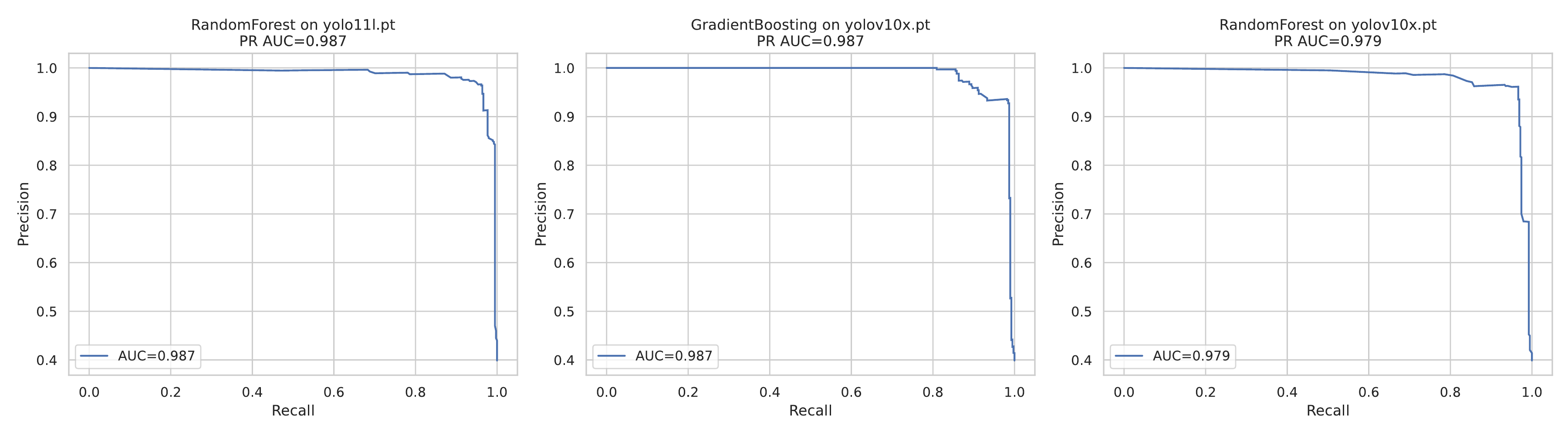}} \\
  \subfloat[\small{Confusion Matrices for top 3 Classifiers}\label{fig:vfpd_confusion}]{\includegraphics[width=0.95\linewidth]{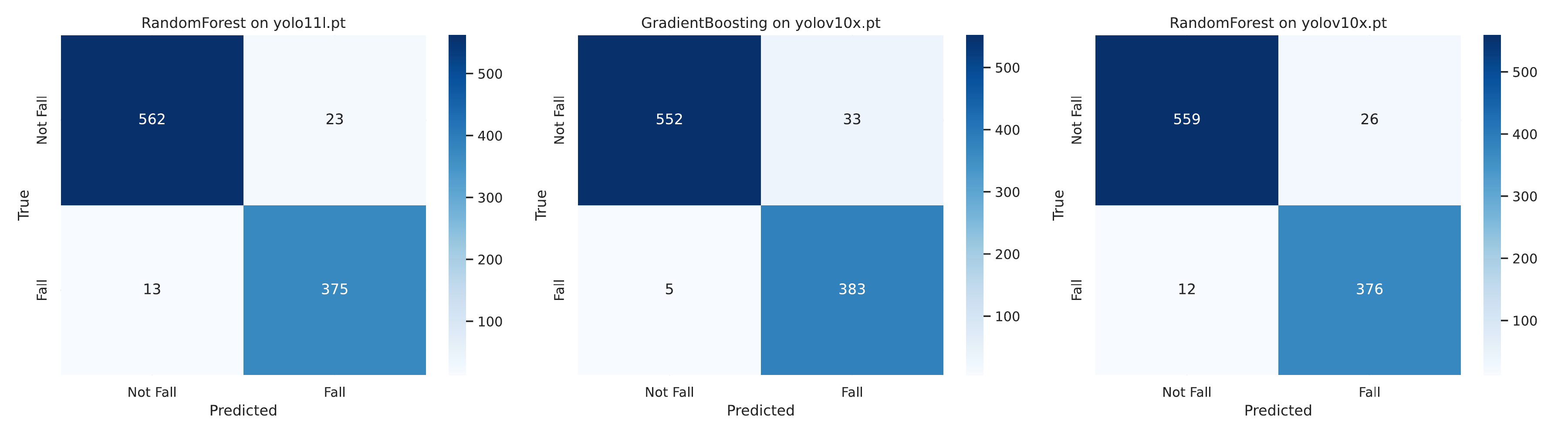}}        
  \caption{Detailed information on the performance of the best classifiers}
    \label{fig:classifiers_results}
\end{figure*}

In Figure~\ref{fig:classifiers_results}, we present the PR (Precision-Recall) curves and AUC (Area Under the Curve) scores for the top 3 classifiers, shown in panel (a). The AUC values clearly demonstrate that all three top combinations have achieved a reliable performance. Both classifiers show strong performance in distinguishing between classes, but \texttt{Random Forest} slightly edges out \texttt{Gradient Boosting} in terms of AUC, although the difference is minimal. Panel (b) of the same figure provides confusion matrices for the top 3 classifiers. From these matrices, we observe that \texttt{Random Forest} has a higher number of correct predictions across all classes, particularly for classifying positive instances. This suggests its superior ability to identify true positives compared to \texttt{Logistic Regression}, which shows more false negatives.

The detailed performance information for the top classifier combinations is shown in Table~\ref{tab:top_3_dombination}. This table presents the results for the top three combinations of YOLO models and classifiers. We see that the combination of \texttt{yolo11l} with \texttt{Random Forest} achieves the highest accuracy at 0.963, along with the best macro and weighted performance metrics. The second-best combination, \texttt{yolov10x} with \texttt{Gradient Boosting}, achieves 0.961 accuracy and shows slightly better recall and F1 scores than the \texttt{yolov10x} and \texttt{Random Forest} combination. All three combinations perform exceptionally well, with strong values across macro and weighted precision, recall, and F1 scores, indicating that these model combinations are effective for the multi-stage detection pipeline.

\begin{table}[ht]
\footnotesize  % Reduce font size to help it fit
\centering
\caption{Performance Comparison of Average YOLO Models Metrics for Various Classifier Models (ACC = Accuracy, PR = Precision, RE = Recall, and F1 = F1 score). \scriptsize{The best result is in \textbf{bold}, and the second best result is \underline{underlined}.}}
\begin{tabular}{lrrrr}
\toprule
\textbf{Classifier} & \textbf{PR}$\uparrow$ & \textbf{RE}$\uparrow$ & \textbf{F1}$\uparrow$  & \textbf{ACC}$\uparrow$ \\
\midrule
RandomForest & 0.909042 & \underline{0.918336} & 0.910297 & 0.912201 \\
LogisticRegression & 0.902340 & 0.880097 & 0.887480 & 0.894435 \\
GradientBoosting & \textbf{0.932014} & \textbf{0.938280} & \textbf{0.933421} & \textbf{0.935325} \\
SVM & \underline{0.910548} & 0.916107 & \underline{0.911640} & \underline{0.914403} \\
\bottomrule
\end{tabular}
\label{tab:classifier_averages}
\end{table}

The overall classifier performance is summarized in Table~\ref{tab:classifier_averages}, which presents average metrics across all YOLO models for each classifier. From the table, we observe that \texttt{Gradient Boosting} achieves the highest scores in terms of precision (0.932), recall (0.938), F1-score (0.933), and accuracy (0.935). On the other hand, \texttt{Logistic Regression} shows slightly lower performance across all metrics, with a notable decrease in recall (0.880) and F1-score (0.887).

The overall performance comparison of average YOLO model metrics across various classifier models is summarized in Table~\ref{tab:classifier_averages}. While the combination of \texttt{Random Forest} performs well across all metrics, with a precision of 0.909 and recall of 0.918, \texttt{Gradient Boosting} outperforms the other classifiers on average, achieving higher precision (0.932), recall (0.938), and F1-score (0.933), along with an accuracy of 0.935. This suggests that although two of the top three combinations of YOLO models and classifiers use \texttt{Random Forest} (as seen in Table~\ref{tab:top_3_dombination}), \texttt{Gradient Boosting} performed better on average across all metrics.

\begin{table*}[ht]
\footnotesize  % Reduce font size to help it fit
\centering
\caption{Detailed information on the performance of the best classifiers. (ACC = Accuracy, PR = Precision, RE = Recall, and F1 = F1 score)}
\begin{tabular}{llrrrrrrr}
\toprule
\textbf{YOLO Model} & \textbf{Classifier} & \textbf{ACC}$\uparrow$ & \textbf{Macro PR}$\uparrow$ & \textbf{Macro RE}$\uparrow$ & \textbf{Macro F1}$\uparrow$ & \textbf{Weighted PR}$\uparrow$ & \textbf{Weighted RE}$\uparrow$ & \textbf{Weighted F1}$\uparrow$ \\
\midrule
yolo11l.pt & RandomForest & 0.963 & 0.960 & 0.964 & 0.962 & 0.963 & 0.963 & 0.963 \\
yolov10x.pt & GradientBoosting & 0.961 & 0.956 & 0.965 & 0.960 & 0.963 & 0.961 & 0.961 \\
yolov10x.pt & RandomForest & 0.961 & 0.957 & 0.962 & 0.960 & 0.962 & 0.961 & 0.961 \\
\bottomrule
\end{tabular}
\label{tab:top_3_dombination}
\end{table*}

\begin{figure}[htb] 
    \centering
    \includegraphics[width=0.95\linewidth]{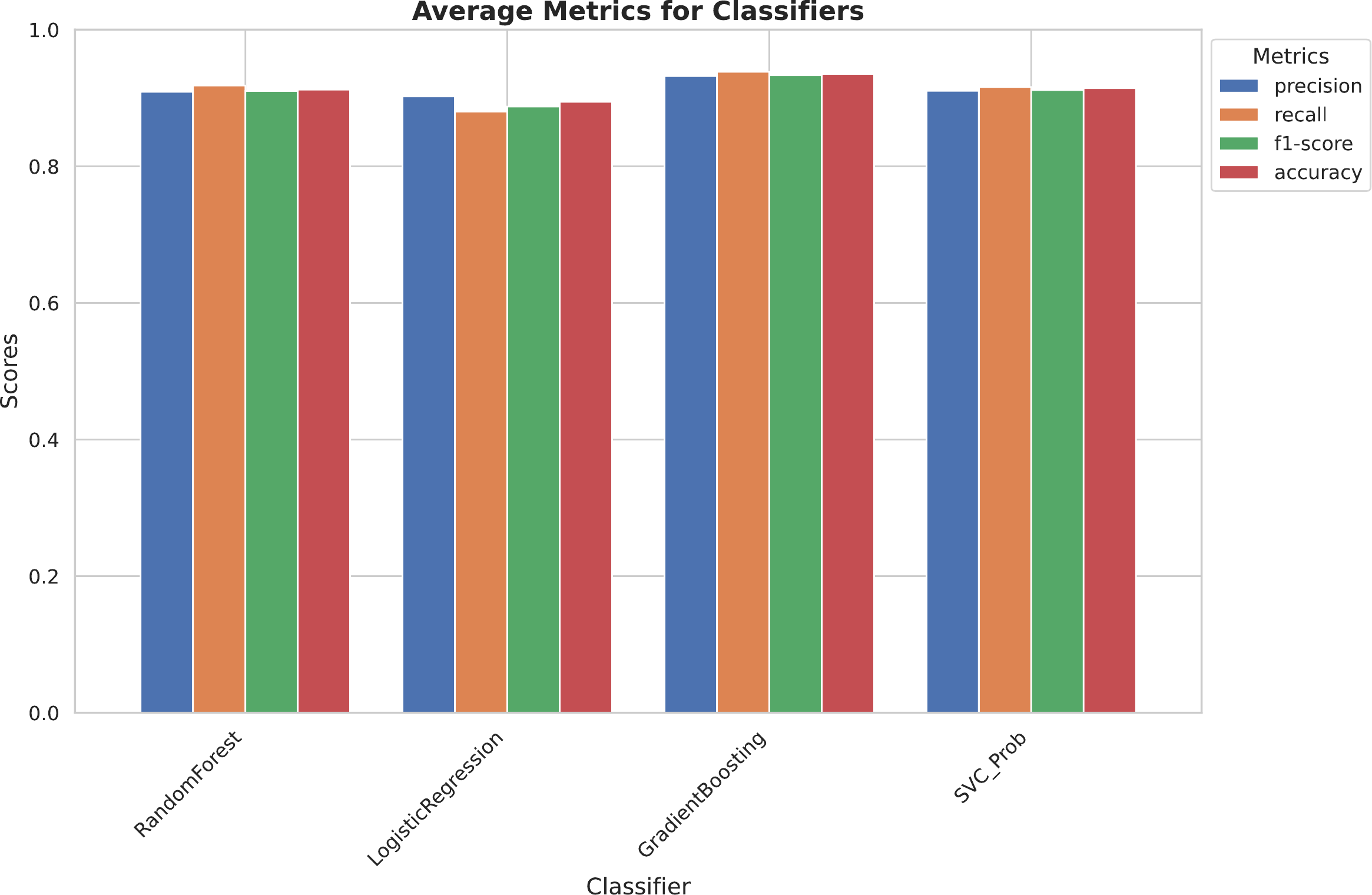}
    \caption{Performance Comparison of Average YOLO Models Metrics For Various Classifier Models}
    \label{fig:plot_classifier_averages}
\end{figure}

\textbf{YOLO Model Performance Comparison:} Figure~\ref{fig:plot_classifier_averages} compares the performance of average YOLO model metrics across various classifier models. This plot reveals that the performance of the YOLO models, particularly \texttt{yolov10x} and \texttt{yolo11x}, consistently improves with the use of more advanced classifiers such as \texttt{Gradient Boosting} and \texttt{Random Forest}, highlighting the synergy between the object detection and classification steps.

\begin{figure}[htb] 
    \centering
    \includegraphics[width=0.95\linewidth]{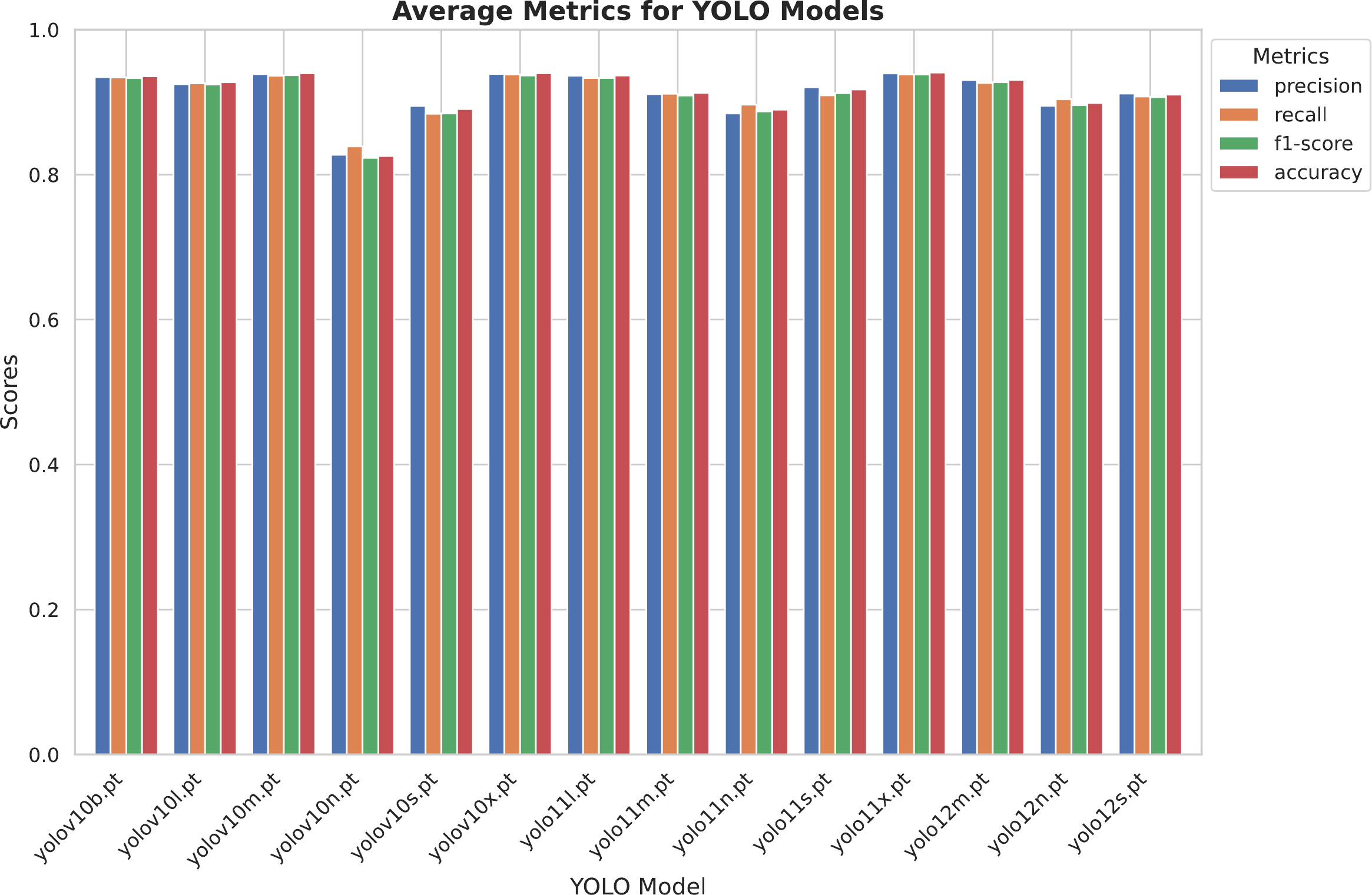}
    \caption{Performance Comparison of Average Classifiers Models Metrics For Various YOLO Models}
    \label{fig:plot_yolo_model_averages}
\end{figure}

Similarly, Figure~\ref{fig:plot_yolo_model_averages} and Table \ref{tab:yolo_model_averages} show the performance comparison of classifiers across various YOLO models. We observe that \texttt{yolov10x} and \texttt{yolo11x} perform the best, achieving the highest metrics, while \texttt{yolov10n} demonstrates relatively lower performance, particularly in precision and recall.

\begin{table}[ht]
\footnotesize  % Reduce font size to help it fit
\centering
\caption{Performance Comparison of Average Classifiers Models Metrics For Various YOLO Models (ACC = Accuracy, PR = Precision, RE = Recall, and F1 = F1 score). \scriptsize{The best result is in \textbf{bold}, and the second best result is \underline{underlined}.}}
\begin{tabular}{lrrrr}
\toprule
YOLO Model & PR$\uparrow$ & RE$\uparrow$ & F1$\uparrow$ & ACC$\uparrow$\\
\midrule
yolov10b & 0.934599 & 0.934213 & 0.933083 & 0.935766 \\
yolov10l & 0.924840 & 0.925965 & 0.924488 & 0.927544 \\
yolov10m & 0.938540 & 0.936330 & \underline{0.937117} & \underline{0.939877} \\
yolov10n & 0.827290 & 0.838855 & 0.823096 & 0.825797 \\
yolov10s & 0.894741 & 0.884022 & 0.884422 & 0.890545 \\
yolov10x & \underline{0.939125} & \textbf{0.938283} & 0.936874 & \underline{0.939877} \\
yolo11l & 0.936639 & 0.933441 & 0.933403 & 0.936793 \\
yolo11m & 0.910979 & 0.911616 & 0.909238 & 0.912898 \\
yolo11n & 0.884406 & 0.896616 & 0.887275 & 0.889774 \\
yolo11s & 0.920648 & 0.909278 & 0.912510 & 0.917523 \\
yolo11x & \textbf{0.939567} & \underline{0.938161} & \textbf{0.938233} & \textbf{0.940904} \\
yolo12m & 0.930592 & 0.926356 & 0.927524 & 0.930884 \\
yolo12n & 0.894938 & 0.903987 & 0.895746 & 0.898767 \\
yolo12s & 0.911903 & 0.907744 & 0.906925 & 0.910329 \\
\bottomrule
\end{tabular}
\label{tab:yolo_model_averages}
\end{table}

\textbf{Trade-Off Between Inference Time and Accuracy}:
There is a clear trade-off between the inference time and the accuracy of YOLO models. Larger models like \texttt{yolov10x} and \texttt{yolo11x}, which achieve high accuracy (around 0.940), come with higher inference times (0.147 and 0.138 seconds, respectively) as shown in Table~\ref{tab:yolo_performance}. In contrast, smaller models such as \texttt{yolov10n} and \texttt{yolo11n}, with lower accuracy (0.826 and 0.890), offer significantly faster inference times (0.027 and 0.028 seconds).

For applications requiring high speed, lighter models may be preferred despite their lower accuracy, while for tasks that prioritize precision, larger models are more suitable, albeit with a trade-off in speed. The choice between speed and accuracy should be based on the specific needs of the application.

\textbf{Discussion:} The results from the various figures and tables consistently show that \texttt{Gradient Boosting} and \texttt{Random Forest} classifiers, when paired with advanced YOLO models such as \texttt{yolov10x} and \texttt{yolo11x}, deliver the highest performance. This suggests that a multi-stage detection pipeline using these models can effectively capture and classify objects in a given scene. Furthermore, the detailed analysis of the confusion matrices and classifier performance metrics offers valuable insights into the strengths and weaknesses of different model combinations for various detection tasks.

\subsection{End-to-End Evaluation}

To assess the real-world applicability and robustness of our system, we conducted an end-to-end experimental deployment in a laboratory setting. This experiment tested the complete pipeline of our framework, starting from IMU-based fall detection to autonomous robot-assisted visual confirmation. The goal was to verify whether the integrated system components, including wearable sensors, machine learning models, edge computing, and robotic navigation, function reliably in a continuous, online scenario.

As described in Section~\ref{sec:method_overview}, the workflow begins with continuous IMU data collection. Upon detecting a fall, the wearable device triggers the localization module and sends a navigation request to the robot. Simultaneously, the robot streams live camera data to the edge, where a YOLO-based visual detector processes frames at 2 FPS. Annotated frames and navigation feedback are displayed in RViz2.

\textbf{Nav2 + SLAM Toolbox Performance:} Our navigation system is powered by the Nav2 stack integrated with SLAM Toolbox for online mapping and localization. According to benchmark reports from the official documentation and developer evaluations \citep{macenski2020marathon, macenski_slam_2021}, the Nav2 stack consistently achieves high navigation reliability with a success rate exceeding 95\% in structured indoor environments. SLAM Toolbox maintains localization accuracy within 2-5 cm under nominal conditions, while the dynamic costmap integration enables replanning latencies between 100-250 ms. These performance characteristics ensure timely and accurate navigation in our fall detection scenarios.

\textbf{Qualitative Observations:} Figure~\ref{fig:rviz_fallen} illustrates a complete successful scenario. The robot navigates to the correct location and detects the fallen person using onboard vision. In contrast, Figure~\ref{fig:rviz_not_fallen} shows a false alarm case where the robot reaches the person, but correctly identifies them as not fallen. Finally, Figure~\ref{fig:rviz_extra} demonstrates two important features: dynamic obstacle avoidance using a live costmap, and successful disambiguation of seated individuals who are commonly misclassified by signal-only systems.

\textbf{Quantitative Evaluation:} Each component of our system contributes to the overall performance, with the success of one component dependent on the preceding one. The combined system performance can be calculated by multiplying the failure rates of each individual system component. 

\begin{itemize}
    \item \textbf{Fall Detection:} 99.19\% accuracy (0.0081 failure rate)
    \item \textbf{Navigation (Nav2 + SLAM Toolbox):} 95\% success rate (0.05 failure rate)
    \item \textbf{Vision-based Fallen People Detection:} 96.3\% accuracy (0.0367 failure rate)
\end{itemize}

The combined failure rate of the system is the product of the failure rates of all components:

\[
\text{Combined Failure Rate} = 0.0081 \times 0.05 \times 0.0367 = 
\]
\[
0.0000149
\]
Thus, the combined accuracy of the system is:

\[
\text{Combined Accuracy} = 1 - \text{Combined Failure Rate} = 
\]
\[
1 - 0.0000149 = 99.99851\%
\]

This calculation highlights the overall high reliability of the system, with a combined accuracy of approximately 99.9985\%.

The end-to-end experiment confirms the practical viability of our framework for real-time fall detection, localization, navigation, and visual confirmation. The system demonstrates high detection accuracy, fast response, and robust handling of false positives. These results support its real-world deployability in elder care, healthcare monitoring, or ambient assisted living environments.

\section{Discussion}
\label{sec:discussion}
We discussed our system’s impact on healthcare and aging-in-place, highlighting its performance, privacy features, and broader applicability. Key limitations and future directions are also outlined to guide responsible and adaptable AI deployment.
% We explain our approach's limitations.

% Talk about the navigation accuracy and performance
% Implications for health care and emergency contact

% How can we use the misclassified data from both detection algorithms
% Using the wrong prediction to enhance the system.

% \color{teal}
\subsection{Implications in Healthcare and Aging-in-Place}
Our system's demonstrated $>$99\% accuracy aligns with professional-grade deployments that achieve 52-70\% reduction in hospitalizations for people aged 65 and above \citep{england2025nationwide}. 
The economic implications extend beyond individual cost savings to healthcare system transformation. The global fall detection market, projected to reach \$19.28 billion by 2028 \citep{fallmarket2030}, reflects growing recognition that proactive monitoring can fundamentally alter care delivery models.

Beyond technical achievements, this work contributes to the broader goal of technology-enabled aging in place. By providing reliable, privacy-conscious fall detection, the system could reduce healthcare costs associated with delayed emergency response while supporting elderly independence. 

Our privacy-preserving FL approach establishes a template for other healthcare monitoring applications where data sensitivity is paramount. 
With the advances in FL for healthcare, we demonstrate that privacy-preserving systems can achieve comparable performance to centralized approaches while maintaining AI compliance stated in GDPR \citep{regulation2016regulation}, EU AI Act \citep{neuwirth2022eu}, Canada's Directive on Automated Decision-Making \citep{board_board_2019}, and Health Canada \citep{canada2025}.
This could accelerate the adoption of AI-based health monitoring systems by addressing privacy concerns that currently limit their deployment, as privacy concerns in fall detection systems can reduce users' willingness to actually adopt such systems \citep{chaudhuri2014fall,igual2013challenges}.

Our framework's multi-modal design, where AI, robotics, and privacy-preserved edge computing are integrated, also enables adaptation to other emergency scenarios, such as medical episodes or home intrusions. 
The integration of edge computing with low latency while processing data locally addresses the growing demand for real-time healthcare monitoring without compromising sensitive health information \citep{singh2021securing}.
More significantly, our framework enables new models of care delivery that bridge the gap between independent living and institutional care. The system's ability to provide predictive insights and immediate emergency response creates opportunities for virtual care teams where healthcare and telehealth providers can monitor multiple patients remotely while ensuring rapid intervention when needed.

\subsection{Limitations, Challenges and Future Works towards Contestable AI Systems}
The current system requires an initial infrastructure setup and assumes indoor environments with reliable robot navigation paths. While the Nav2 stack demonstrates high reliability ($>$95\% success rate) of our online mapping and localization capabilities, sufficient for the structured indoor-level navigation, it may face challenges in highly cluttered or dynamically changing environmental factors, such as moved furniture, temporary obstacles, or seasonal lighting changes, which could impact performance. Future deployments should incorporate adaptive mapping capabilities to maintain long-term reliability.

The successful handling of false alarms demonstrates the system's ability to learn from misclassifications, potentially leveraging misclassified data for continuous learning. When the robot confirms a false alarm, this information can be fed back to retrain the initial fall detection model, creating a self-improving system. 
Furthermore, our current framework's multi-stage verification process inherently provides a foundation for contestable AI design through its built-in safeguard mechanism. 
The robot-assisted visual verification serves as a secondary automated system that prevents unilateral decisions by the primary fall detection algorithm. This self-adversarial decision-making process, where the wearable sensor's initial classification must be confirmed by an independent robotic verification system, exemplifies the type of procedural safeguard essential for contestable AI systems \citep{alfrink2023contestable,nguyen2025heart2mind,lyons2021conceptualising}.
When inconsistencies arise between the IMU-based detection and visual confirmation, the system can automatically escalate the decision through built-in conflict resolution protocols rather than proceeding with potentially erroneous emergency responses.

In summary, as healthcare AI regulations continue to evolve, demonstrated approaches that balance innovation with privacy protection and human agency could influence policy development and establish best practices for responsible AI deployment in the healthcare context.

% - Hung Jr.
% \color{black}
\section{Conclusion and Future Work}
\label{sec:conclusion}
In conclusion, our proposed fall detection framework presents a novel, highly dependable solution for addressing the growing risk of falls among the aging population. We developed and evaluated a multi-stage fall detection system with the main goals of reducing false alarms and protecting user privacy. By integrating two complementary systems, i.e., a semi-supervised federated learning-based fall detection system and a vision-based fallen person detection technique, the framework ensures timely, reliable fall detection while maintaining privacy. We leveraged the potential of federated learning to protect users' privacy and the power of a mobile vision-based system to enhance reliability without compromising privacy. Together, they achieved a highly accurate overall detection with 99.99\% accuracy, a step toward the aging-in-place paradigm. The future direction for this work can be focused on long-term experiments on older people to discover and address the real-world deployment challenges. Also, research on the optimization and energy efficiency of the edge and wearable devices can contribute to a more reliable system. 

% \section*{Acknowledgments}

\bibliographystyle{elsarticle-harv}
\bibliography{references}

\vfill

\end{document}